\documentclass{svproc}
\usepackage{url}

\usepackage[T1]{fontenc}
\def\doi#1{\href{https://doi.org/\detokenize{#1}}{\url{https://doi.org/\detokenize{#1}}}}
\usepackage{cite}
\usepackage{amsmath,amssymb,amsfonts}
\usepackage{algorithmic}
\usepackage{graphicx}
\usepackage{textcomp}
\usepackage{xcolor}
\usepackage{caption}
\usepackage{subcaption}
\usepackage{academicons}
\usepackage{bbding}
\definecolor{orcidlogocol}{HTML}{A6CE39}

\usepackage{listings}
\usepackage[linesnumbered,ruled,vlined,noend]{algorithm2e}
\usepackage[font=tiny]{caption}
\newcommand{\prob}[1]{\ensuremath{\mathbb{P}({#1})}}

\newcommand{\simplified}{\textit{simplified}}
\newcommand{\budget}{$\mathcal{C}$\xspace}
\newcommand{\skeleton}{\textit{skeleton}\xspace}

\begin{document}
\mainmatter              % start of a contribution
\title{Nonmyopic Distilled Data Association Belief Space Planning Under Budget Constraints\thanks{This work was partially supported by US NSF/US-Israel BSF.}}
\titlerunning{Nonmyopic Distilled Data Association BSP Under Budget Constraints}  % abbreviated title (for running head)
\author{Moshe Shienman \and Vadim Indelman}

\authorrunning{M. Shienman and V. Indelman} % abbreviated author list (for running head)
%Craig Chambers, Kim B. Bruce, and Elisa Bertino}
%
\institute{Technion - Israel Institute of Technology, Haifa 32000, Israel \\
\email{smoshe@campus.technion.ac.il , vadim.indelman@technion.ac.il}}

\maketitle              % typeset the title of the contribution

% abstract
\begin{abstract}
%The abstract should summarize the contents of the paper
%using at least 70 and at most 150 words. It will be set in 9-point
%font size and be inset 1.0 cm from the right and left margins.
%There will be two blank lines before and after the Abstract. \dots
% We would like to encourage you to list your keywords within
% the abstract section using the \keywords{...} command.
%\keywords{computational geometry, graph theory, Hamilton cycles}	
Autonomous agents operating in perceptually aliased environments should ideally be able to solve the data association problem. Yet, planning for future actions while considering this problem is not trivial. State of the art approaches therefore use multi-modal hypotheses to represent the states of the agent and of the environment. However, explicitly considering all possible data associations, the number of hypotheses grows exponentially with the planning horizon. As such, the corresponding Belief Space Planning problem quickly becomes unsolvable. Moreover, under hard computational budget constraints, some non-negligible hypotheses must eventually be pruned in both planning and inference. Nevertheless, the two processes are generally treated separately and the effect of budget constraints in one process over the other was barely studied. We present a computationally efficient method to solve the nonmyopic Belief Space Planning problem while reasoning about data association. Moreover, we rigorously analyze the effects of budget constraints in both inference and planning.

\keywords{Planning under Uncertainty, Robust Perception, SLAM}

\end{abstract}

% Introduction
\section{Introduction}
Intelligent autonomous agents and robots are expected to operate reliably and efficiently under different sources of uncertainty. There are various possible reasons for such uncertainty, including noisy measurements; imprecise actions; and dynamic environments in which some events are unpredictable. In these settings, autonomous agents are required to reason over high-dimensional probabilistic states known as beliefs. A truly autonomous agent should be able to perform both inference, i.e. maintain a belief over the high-dimensional state space given available information, and decision making under uncertainty. The latter is also known as the Belief Space Planning (BSP) problem, where the agent should autonomously determine its next best actions while reasoning about future belief evolution. However, both inference and BSP are computationally expensive and practically infeasible in real-world autonomous systems where the agent is required to operate in real time using inexpensive hardware.

In real-world scenarios, an autonomous agent should also be resilient to the problem of ambiguous measurements. These ambiguities occur when a certain observation has more than one possible interpretation. Some examples include the slip/grip behavior of odometry measurements; the loop closure problem in visual Simultaneous Localization and Mapping (SLAM); and unresolved data association. The latter is defined as the process of associating uncertain measurements to known tracks, e.g. determine if an observation corresponds to a specific landmark within a given map. Most existing inference and BSP algorithms assume data association to be given and perfect, i.e. assume a single hypothesis represented by a uni-modal state and map estimates. Yet, in perceptually aliased environments, this assumption is not reasonable and could lead to catastrophic results. %For example, an autonomous agent might incorrectly infer its location, which can lead to false action selection and possibly a full system failure. 
Therefore, it is crucial to reason about data association, in both inference and planning, while also considering other sources of uncertainty. 

Explicitly reasoning about data association, the number of hypotheses grows exponentially with time. As such, when considering real time operation using inexpensive hardware, hard computational constraints are often required, e.g. bounding the number of supported hypotheses. State of the art inference and planning approaches therefore use different heuristics, e.g. pruning and merging, to relax the computational complexity. However, this loss of information incurs loss in solution quality and there are usually no performance guarantees. Moreover, inference and planning are commonly treated separately and it is unclear how budget constraints in one process affect another.

In this work we extend our presented approach in \cite{Shienman22icra} to a nonmyopic setting. Specifically, we handle the exponential growth of hypotheses in BSP by solving a \simplified \! problem while providing performance guarantees. 
To that end, we analyze for the first time, the construction of a belief tree within planning given a mixture belief, e.g. Gaussian Mixture Models (GMM). We further show how to utilize the \skeleton \! of such belief tree to reduce the computational complexity in BSP.
Crucially, this paper thoroughly studies, for the first time, the impacts of hard budget constraints in either planning  and/or inference.

% Related Work
\section{Related Work}
Several approaches were recently proposed to ensure efficient and reliable operation in ambiguous environments. Known as robust perception, these approaches typically maintain probabilistic data association and hypothesis tracking.

A good inference mechanism should handle false data association made by front-end algorithms and be computationally efficient. The authors of \cite{Shelly22ral} recently suggested to re-use hypotheses’ weights from previous steps to reduce computational complexity and improve current-time hypotheses pruning. Convex relaxation approaches over graphs were proposed in \cite{Carlone14iros, Lajoie19ral} to capture perceptual aliasing and find the maximal subset of internally coherent measurements, i.e. correct data association. The max-mixture model was presented in \cite{Olson13ijrr} to allow fast maximum-likelihood inference on factor graphs \cite{Kschischang01it} that contain arbitrarily complex probability distributions such as the slip/grip multi modal problem. The authors of \cite{Indelman14icra, Indelman16csm} used factor graphs with an expectation-maximization approach to efficiently infer initial relative poses and solve a multi robot data association problem. In \cite{Sunderhauf12icra} the topological structure of a factor graph was modified during optimization to discard false positive loop closures. The Bayes tree algorithm \cite{Kaess12ijrr} was extended in \cite{Hsiao19icra, Jiang21arxiv_b} to explicitly incorporate multi-modal measurements within the graph and generate multi-hypothesis outputs. These works, however, were only developed for the purpose of inference, i.e. without planning.

Ambiguous data association was also considered in planning. In \cite{Agarwal16wafr} a GMM was used to model prior beliefs representing different data association hypotheses. However, the authors did not reason about ambiguous data association within future beliefs (owing to future observations), i.e. they assumed that it is solved and perfect in planning. In \cite{Pathak18ijrr} the authors introduced DA-BSP where, for the first time, reasoning about future data association hypotheses was incorporated within a BSP framework. The ARAS framework proposed in  \cite{Hsiao20iros} leveraged the graphical model presented in \cite{Hsiao19icra} to reason about ambiguous data association in future beliefs. All of these approaches handled the exponential growth in the number of hypotheses by either pruning or merging. The first work to also provide performance guarantees on the loss in solution quality was presented in \cite{Shienman22icra}. Yet, the authors only considered a myopic setting.  

The notion of simplification was introduced in \cite{Elimelech21ijrr}, where, the authors formulated the loss in solution quality in BSP problems via bounds over the objective function. However, they only considered the Gaussian case and a maximum likelihood assumption. The authors of \cite{Sztyglic21arxiv} used bounds as a function of \simplified \! beliefs to reduce the computational complexity in nonmyopic BSP problems with general belief distributions. In \cite{Sztyglic21arxiv_b} they incorporated this concept within a Monte Carlo Tree Search (MCTS) planning framework, i.e. without assuming that the belief tree is given, which is complimentary to our approach. Yet, they did not handle ambiguous data association nor budget constraints aspects.

%\begin{figure*} [h]
%	\begin{subfigure}{0.5\textwidth}
%		\centering
%		\includegraphics[scale=0.10]{./images/data_association.eps}
%		\caption{unresolved data association}
%		\label{fig: unresolved data association}
%	\end{subfigure}
%	\begin{subfigure}{0.6\textwidth}
%		\centering
%		\includegraphics[scale=0.10]{./images/belief_mixture.eps}
%		\caption{posterior belief}
%		\label{fig: posterior belief}
%	\end{subfigure}
%	\caption{\MS{add caption}}
%	\label{fig: unresolved data association with posterior belief}
%\end{figure*} 

% noatations
\section{Background and Notations}
\label{notations section}
In this section we review some basic concepts from estimation theory and BSP which we will use in the following sections.

\subsection{Inference}
Consider an autonomous agent operating in a partially known or pre-mapped environment
containing similar landmarks or scenes.  The agent acquires observations and tries to infer random variables of interest that are application dependent while reasoning about data association.

We denote the agent's state at time instant $k$ by $x_k$. Let $Z_k \triangleq \lbrace z_{k,1},...,z_{k,n_k} \rbrace$ denote the set of all $n_k$ measurements and let $u_k$ denote the agent's action. $Z_{1:k}$ and $u_{0:k-1}$ denote all observations and actions up to time $k$, respectively. The motion and observation models are given by
\begin{equation} \label{eq:motion and observation models}
	x_{k+1} = f \left( x_k,u_k,w_k \right) \quad , \quad z_{k}=h \left( x_k, x^l, v_k \right),
\end{equation}
where $x^l$ is a landmark pose and $w_k$ and $v_k$ are noise terms, sampled from known motion and measurement distributions, respectively.

Given $n_k$ observations, the data association realization vector is denoted by $\beta_k \in \mathbb{N}^{n_k}$. Elements in $\beta_k$ are associated according to the given observation model and each element, e.g. landmark, is given a unique label. A specific data association hypothesis is thus given by a specific set $j$ of associations up to and including time $k$ and is denoted as $\beta_{1:k}^j$.

At each time step the agent maintains a posterior belief over both continuous and discrete variables given by 
\begin{equation} \label{eq:joint belief}
	b \left[ x_k ,\beta_{1:k} \right] \triangleq \mathbb{P} \left( x_k , \beta_{1:k}|z_{0:k}, u_{0:k-1}\right) = \mathbb{P} \left( x_k, \beta_{1:k} |H_k\right) ,
\end{equation}
where $H_k \triangleq \lbrace Z_{1:k}, u_{0:k-1} \rbrace$ represents history. Using the chain rule, the belief becomes a mixture and can be written as a linear combination of $\vert M_k  \vert$ hypotheses 
\begin{equation} \label{belief mixture}
	b_k  = \sum_{j \in M_k} 
	\underbrace{\mathbb{P} \left( x_k|\beta_{1:k}^j, H_k\right)}_{b^j_k} \underbrace{\mathbb{P} \left( \beta_{1:k}^j | H_k\right)}_{w_k^j},
\end{equation}
where $b^j_k$ is a conditional belief, with some general distribution, and $w_k^j$ is the associated weight. Therefore, $M_k$ is a set of maintained weighted conditional beliefs, representing different data association hypotheses. In this work, we interchangeably refer to each $b^j_k$ as both a hypothesis and a component. 

Each conditional belief hypothesis $b^j_k$ in \eqref{belief mixture} can be efficiently calculated by maximum a posteriori inference, e.g. as presented in \cite{Kaess12ijrr} for the Gaussian case. Nevertheless, our formulation and approach also applies to a non-parametric setting. 
Each component weight $w_k^j$ is calculating by marginalizing over the state space and applying the Bayes rule (as developed in \cite{Pathak18ijrr, Shienman22icra}).

Reasoning about data association, without any computational constraints, the number of considered hypotheses grows exponentially with time. In general, such belief is a function of $b_k = \psi_k \left( b_{k-1}, u_{k-1}, Z_k \right)$. However, under hard computational constraints, the number of hypotheses is bounded by $\mathcal{C} \in \mathbb{N}$. Therefore, the belief in each time step is a function of
\begin{equation} \label{eq: inference psi}
	b_k^\psi = \psi_k^{\mathcal{C}} \left( b_{k-1}, u_{k-1}, Z_k, {\mathcal{C}} \right),
\end{equation}
where $\psi_k^{\mathcal{C}}$ contains some heuristic function $h^{inf}$ such that $ | M_k^\psi | \leq \mathcal{C}$. 

\begin{figure*} [h]
	\begin{subfigure}{0.5\textwidth}
		\centering
		\includegraphics[scale=0.20]{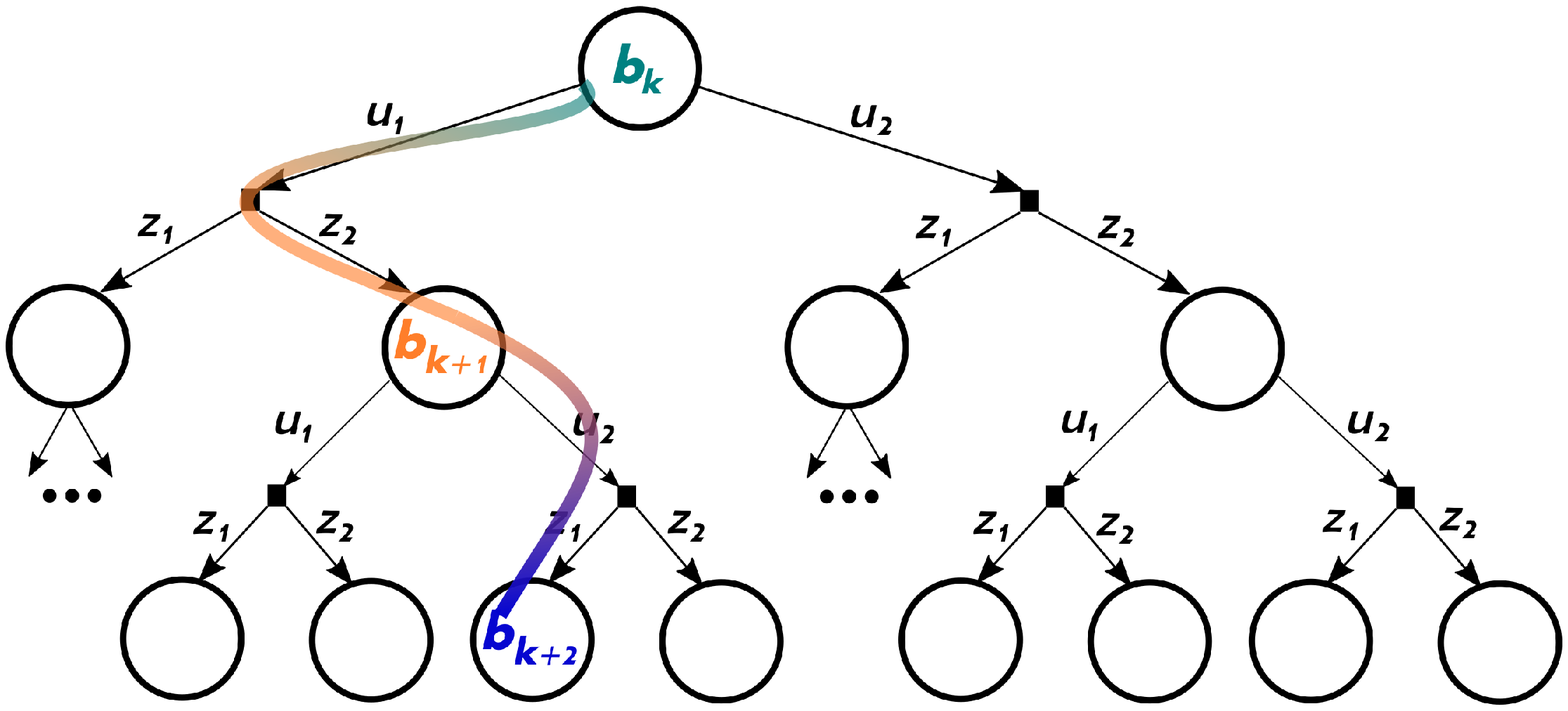}
		\caption{belief tree}
		\label{fig: belief tree}
	\end{subfigure}
	\begin{subfigure}{0.6\textwidth}
		\centering
		\includegraphics[scale=0.20]{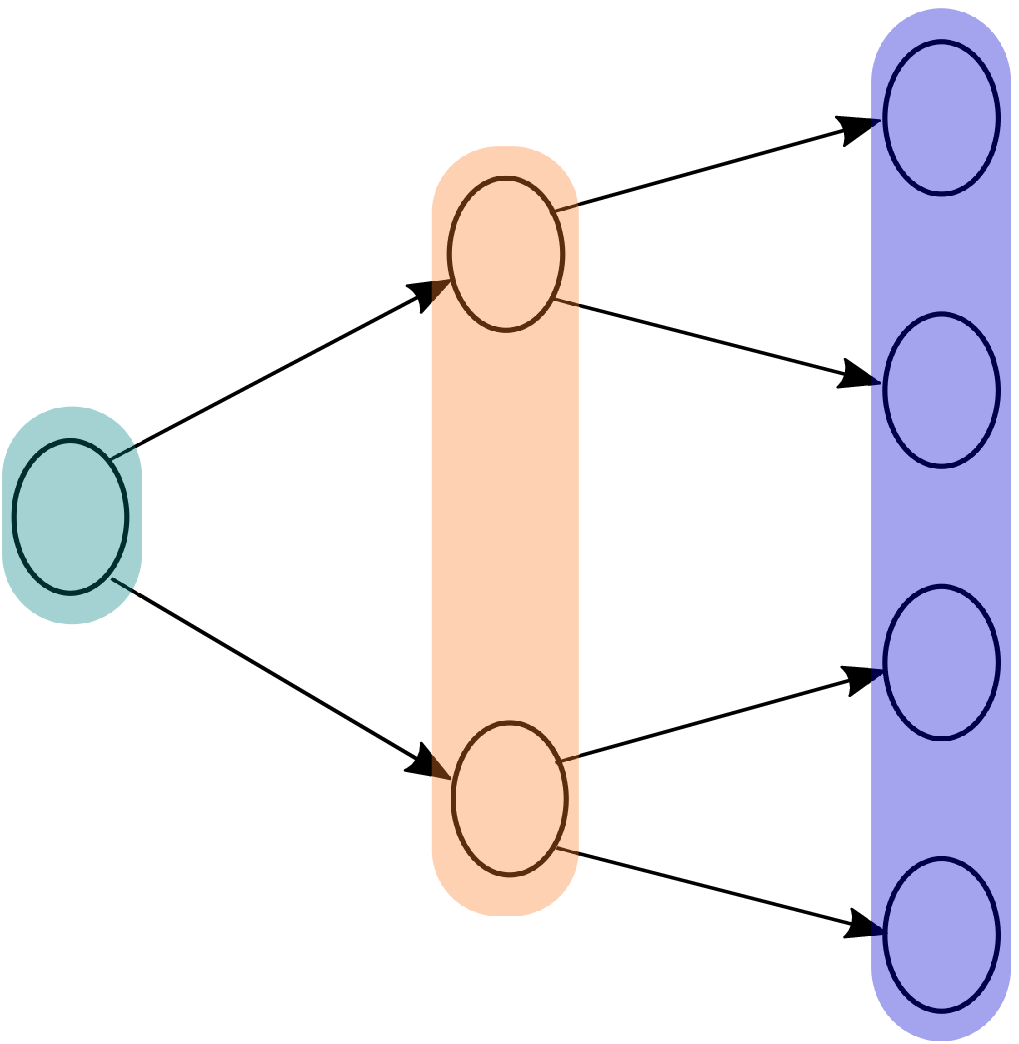}
		\caption{exponential growth of hypotheses}
		\label{fig:exponential growth}
	\end{subfigure}
	\caption{(a) A belief tree constructed during planning. Each node represents a posterior belief \eqref{belief mixture}; The number of belief components grows exponentially along the highlighted path as presented in (b).}
	\label{fig: belief tree with exponential growth}
\end{figure*} 

\subsection{DA-BSP} \label{fomulation bsp}
Given a posterior belief \eqref{belief mixture} and a set of candidate action sequences $\mathcal{U}$ the goal of BSP is to find the optimal action sequence that would minimize/maximize a certain objective function. We note that while in this paper we consider, for simplicity, action sequences, our approach is applicable also to policies.

Reasoning about data association in planning, a user defined objective function $J$ can be written as \begin{equation} \label{eq:general objective}
	J \left(b_k, u_{k:k+N-1} \right)  =  \underset{\beta_{(k+1)+}}{\mathbb{E}} \; \left[ \underset{Z_{(k+1)+} | \beta_{(k+1)+}}{\mathbb{E}} \left[ \sum_{n=1}^{N}   c \left( b_{k+n}, u_{k+n-1} \right) \right] \right],
\end{equation}
where $\beta_{(k+1)+} \triangleq \beta_{k+1:k+N}$ ,  $Z_{(k+1)+} \triangleq Z_{k+1:k+N}$ and 
$c \left( \cdot \right)$ denotes a cost function. The expectation is taken with respect to both future data association realizations and observations.
The optimal action sequence \small $u_{k:k+N-1}^*$ \normalsize is defined as
\begin{equation} \label{optimization objective}
	u_{k:k+N-1}^* = \underset{\mathcal{U}}{\text{argmin}} \, J \left( b_k, u_{k:k+N-1} \right).
\end{equation}
To solve \eqref{optimization objective} we need to consider all possible future realizations of $Z_{k+n}$ for every $n \in \left[k+1,k+N \right]$ while 
marginalizing over all possible locations and data association realizations (see Section 5.2 in \cite{Pathak18ijrr}). However, solving these integrals analytically is typically not feasible. %\MS{consider if required We denote this problem as the \theoretical problem.} 
In practice, the solution should be approximated by sampling future observations from the relevant distributions. Using these samples, the agent constructs and traverses a belief tree (as shown in Fig. \ref{fig: belief tree}) which branches according to future actions and observations. 

Nevertheless, the number of hypotheses grows exponentially with the planning horizon (see Fig. \ref{fig:exponential growth}). Specifically, given $\left| M_k \right|$ hypotheses and $D$ data association realizations, i.e. different $\beta_{k+i}$ at each look-ahead step, the number of belief components at the $n$th look-ahead step is $ \left| M_{k+n} \right|=  \left| M_k \right| \left| D \right|^n$. As such, considering every possible future hypothesis is not practical.

% approach
\section{Methodology} \label{sec: methods}
In this section we first describe how to construct a belief tree \skeleton during planning. We then present a general framework to reduce the computational complexity when solving a sampling based approximation of \eqref{eq:general objective}. Finally, we analyze the implications of using our proposed framework under different conditions. 

\subsection{Constructing the belief tree \skeleton} \label{sec: construct belief tree}
Previous works addressed the exponential growth of the belief tree with the planning horizon without reasoning about data association. In this work we analyze and describe, for the first time, the structure of a belief tree given a mixture belief such as \eqref{belief mixture}. In this setting there is an additional exponential growth in the number of belief components for every considered future observation realization (see Fig. \ref{fig: belief tree with exponential growth}). These realizations are functions of future beliefs \eqref{belief mixture}, data association realizations and actions
\begin{equation} \label{eq:likelihood distributions}
	\prob{Z_{k+1:k+n} | b_k, u_{k:k+n-1}, \beta_{k+1:k+n}}.
\end{equation}
To construct the belief tree in practice, we sample states from beliefs, sample data association given states and finally sample observations from \eqref{eq:likelihood distributions}.

Our key observation is that in order to construct a belief tree \skeleton, i.e. without explicitly calculating or holding posterior beliefs at each node, we can sample future observations in two different ways. We describe these two options for a planning horizon of $n=2$. 
Specifically, we can either rewrite \eqref{eq:likelihood distributions} as
\begin{equation} \label{eq: sample from posterior}
	\prob{Z_{k+2} | b_{k+1|\beta_{k+1}}, u_{k+1}, \beta_{k+2}} \prob{Z_{k+1} | b_{k}, u_{k}, \beta_{k+1}},
\end{equation}
where $ b_{k+1|\beta_{k+1}}$ is a posterior belief and each term is evaluated by integrating over $x_{k+1:k+2}$, or, by first integrating and then applying the chain rule as 
\begin{equation} \label{eq: sample from propagated}
	\int\displaylimits_{x_{k+2}} \!\!
	\prob{Z_{k+2} | x_{k+2}, \beta_{k+2}} \!\!
	\int\displaylimits_{x_{k+1}} \!\!
	\prob{x_{k+2} | x_{k+1}, u_{k+1}} \prob{Z_{k+1} | x_{k+1}, \beta_{k+1}}
	\prob{x_{k+1} | b_{k}, u_k}.
\end{equation}
While these two expressions are analytically identical, they represent two different processes of sampling. In the former observations are sampled from posterior beliefs, while in the latter observations are sampled using the motion and observation models, similar to the MCTS particle trajectories techniques in \cite{Silver10nips, Ye17jair}. 
\vspace{-8pt}
\begin{algorithm}
	\smaller
	\SetAlgoLined
	\KwIn{prior belief $b_k$, action sequence $u_{k:k+n-1}$}
	\KwOut{sampled future observations $Z_{k+1:k+n}$}	
	$Z = \emptyset$ \\
	$x_k \sim b_k$ \\
	\For{$i \in \left[1, n\right]$} {
		$x_{k+i} \sim \prob{x_{k+i} | x_{k+i-1}, u_{k+i-1}}$ \\
		determine $\beta_{k+i}$ based on  $x_{k+i}$ \\
		$Z_{k+i} \sim \prob{Z_{k+i} |x_{k+i}, \beta_{k+i}}$ \\
		$Z = Z \cup Z_{k+i}$}
	\Return $Z$
	\caption{\smaller Construct belief tree \skeleton}
	\label{alg: sample observations}
\end{algorithm}
\vspace{-8pt}

To avoid the explicit representation of the exponential number of belief components, in this work we sample future observations using \eqref{eq: sample from propagated} and bypass the inference stage. We formulate this sampling method in Algorithm \ref{alg: sample observations}.

Yet, this is of little help if the posterior belief is required for calculating the cost function itself. We next describe our approach to avoid these calculations.

\subsection{Nonmyopic Distilled Data Association BSP}
Our goal is to reduce the computational complexity of nonmyopic BSP problems where ambiguous data association is explicitly considered, i.e. solving \eqref{optimization objective} efficiently. We start by writing \eqref{eq:general objective} in a recursive form
\begin{equation} \label{eq:general objective recursive}
	J \left(b_k, u_{k:k+N-1} \right)  = c \left( b_k, u_k \right) +  \underset{\beta_{k+1}}{\mathbb{E}} \left[ \underset{Z_{k+1} | \beta_{k+1}}{\mathbb{E}} \left[ J \left(b_{k+1}, u_{k+1:k+N-1} \right) \right] \right].
\end{equation}
As in practice we approximate the solution via samples, we rewrite \eqref{eq:general objective recursive} as
\begin{equation} \label{eq:emprical objective recursive}
	\hat{J} \left(b_k, u_{k:k+N-1} \right)  = c \left( b_k, u_k \right) +  \underset{\beta_{k+1}}{\hat{\mathbb{E}}} \left[ \underset{Z_{k+1} | \beta_{k+1}}{\hat{\mathbb{E}}} \left[ \hat{J} \left(b_{k+1}, u_{k+1:k+N-1} \right) \right] \right].
\end{equation}
Using Bellman's principle of optimality, the optimal solution for \eqref{eq:emprical objective recursive} is
\begin{equation} \label{eq: bellman objective approx}
	\hat{J} \left(b_k, \hat{u}_{k:k+N-1}^* \right)  = \underset{u_k}{\text{min}} \{ c \left( b_k, u_k \right) +  \underset{\beta_{k+1}}{\hat{\mathbb{E}}} \left[ \underset{Z_{k+1}|\beta_{k+1}}{\hat{\mathbb{E}}} \left[ \hat{J} \left(b_{k+1}, u_{k+1:k+N-1}^* \right) \right] \right] \},
\end{equation}
where \small $\hat{u}_{k:k+N-1}^* = \underset{\mathcal{U}}{\text{argmin}} \, \hat{J} \left( b_k, u_{k:k+N-1} \right)$. \normalsize
To reduce the computational complexity in \eqref{eq: bellman objective approx}, we propose utilizing the belief tree \skeleton, without having access to posterior beliefs, to solve an easier to compute version of the considered cost function.
\begin{figure*} [h]
	\vspace{-10pt}
	\begin{subfigure}{0.5\textwidth}
		\centering
		\includegraphics[scale=0.25]{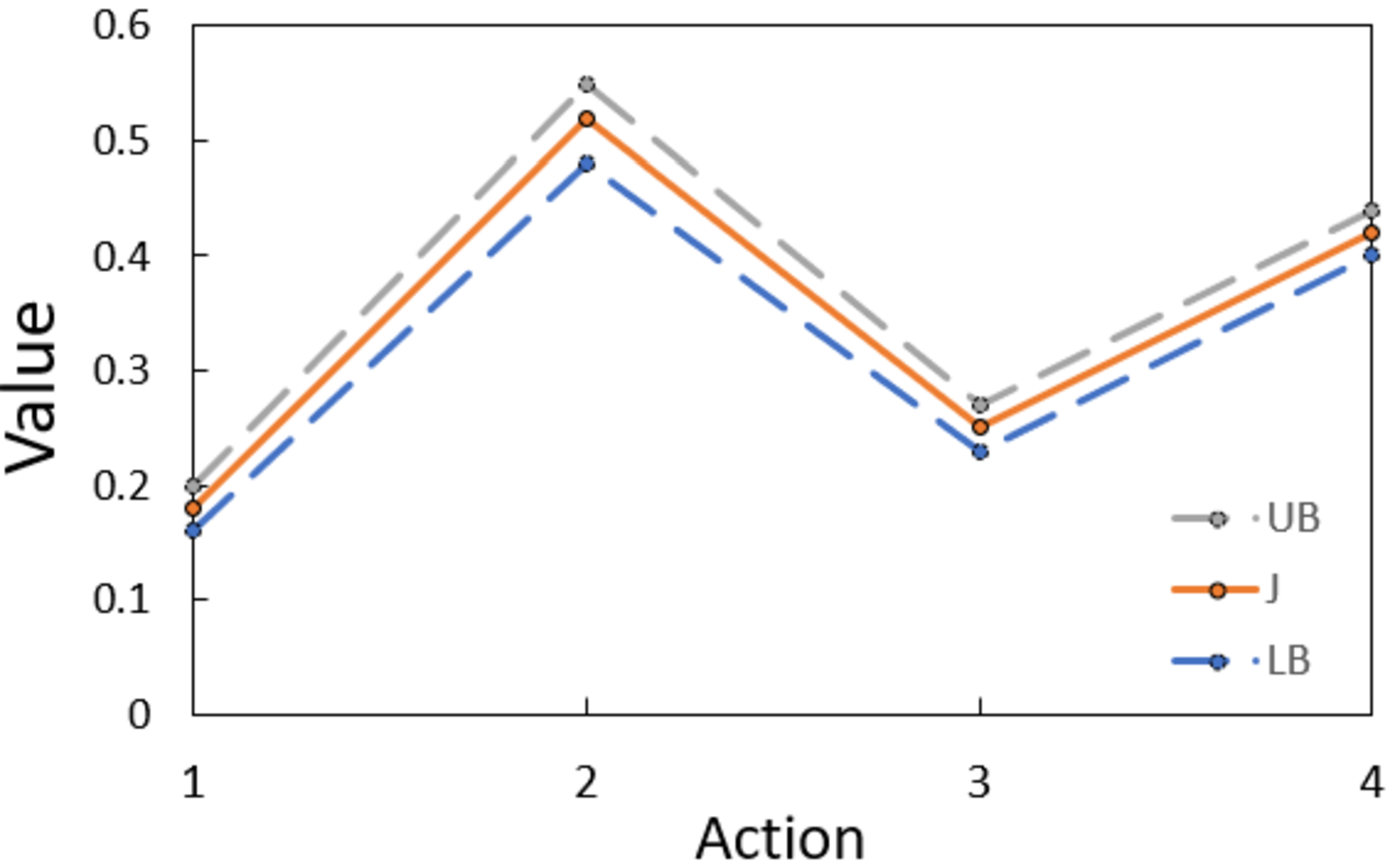}
		\caption{No overlap between bounds}
		\label{fig: bounds no overlap}
	\end{subfigure}
	\begin{subfigure}{0.5\textwidth}
		\centering
		\includegraphics[scale=0.25]{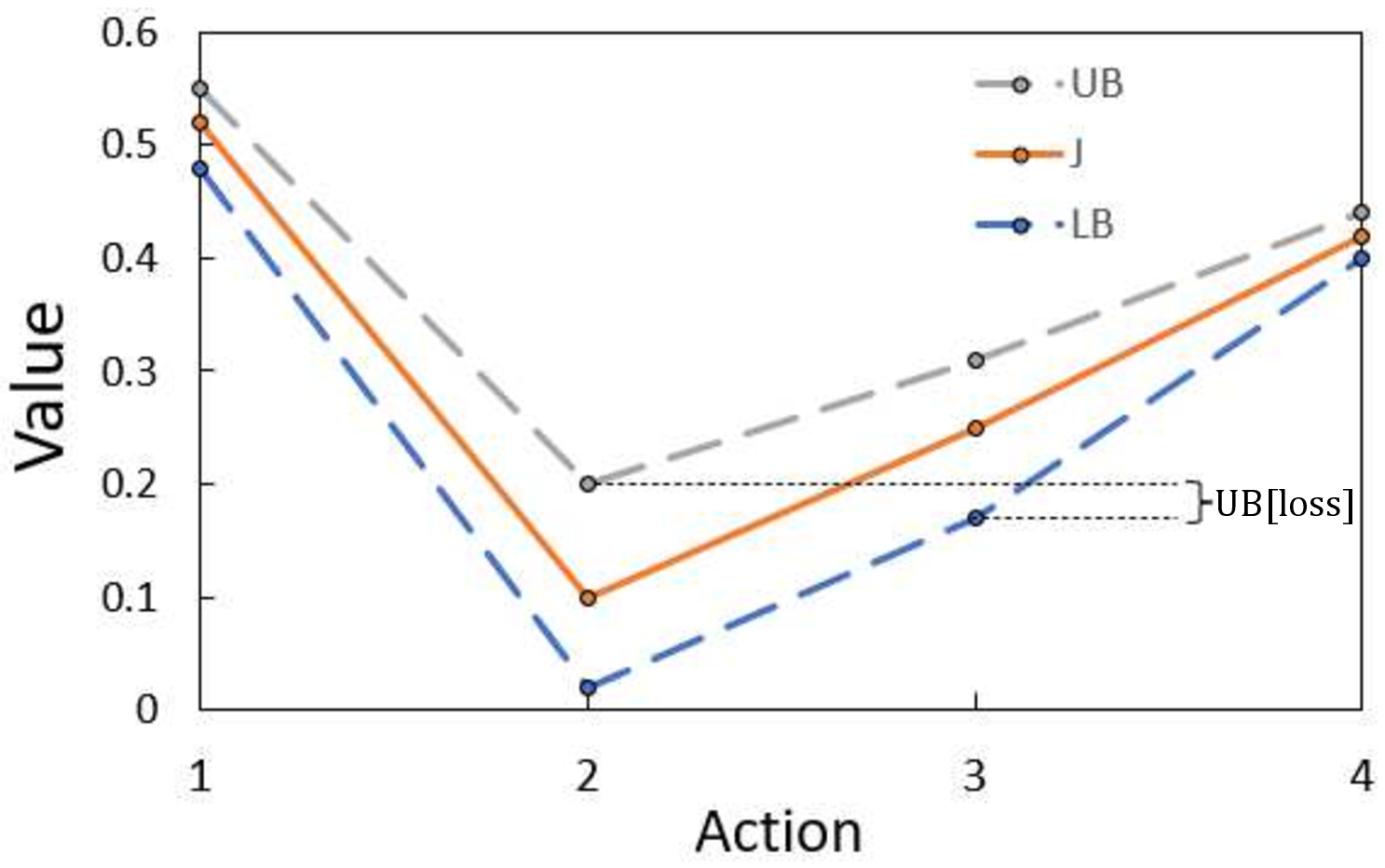}
		\caption{Bounds overlap}
		\label{fig: bounds overlap}
	\end{subfigure}
	\caption{BSP using bounds over the objective function. In (a) choosing action \#1 is guaranteed to be optimal as the corresponding upper bound is lower than all other lower bounds; In (b) choosing action \#2 is not guaranteed to be optimal. The loss in solution quality, however, is upper bounded.}
	\label{fig: planning with bounds}
	\vspace{-16pt}
\end{figure*}
In general, the cost function over the original beliefs can be bounded using a \simplified \! \! belief $b_{k}^s$ as
\begin{equation} \label{eq: general cost bounds}
	\underline{c} \left( b_k^s, u_k \right) \leq c \left( b_{k} , u_k \right) \leq \bar{c} \left( b_k^s, u_k \right).
\end{equation}
We note that this formulation also supports replacing the cost function itself with a computationally simpler function, as in \cite{Kitanov18icra, Shienman21ral}.

Using the belief tree \skeleton and some method to calculate the \simplified \! \! beliefs, to be defined, we now traverse the belief tree from the leafs upwards. At each belief tree node the bounds over the objective function \eqref{eq:general objective} are calculated recursively using the Bellman equation \eqref{eq: bellman objective approx} and \eqref{eq: general cost bounds} for every $n \in \left[ 0, N - 1 \right]$
\begin{equation*} 
	\underline{J} \left(b_{k+n}, u_{(k+n)+} \right)  = 
	\underline{c} \left( b_{k+n}^s , u_{k+n} \right) +
	\underset{\beta_{k+1}}{\hat{\mathbb{E}}} \!\! \left[ \underset{Z_{k+n+1}|\beta_{k+1}}{\hat{\mathbb{E}}} \left[ \underline{J} \left(b_{k+n+1}, u_{(k+n)+} \right) \right] \right],
\end{equation*}
\begin{equation} \label{eq: recrusive objective bounds}
	\bar{J} \left(b_{k+n}, u_{(k+n)+} \right)  = \bar{c} \left( b_{k+n}^s, u_{k+n} \right) + \underset{\beta_{k+1}}{\hat{\mathbb{E}}} \!\! \left[ \underset{Z_{k+n+1}|\beta_{k+1}}{\hat{\mathbb{E}}} \left[ \bar{J} \left(b_{k+n+1}, u_{(k+n)+} \right) \right] \right],
\end{equation}
where $u_{(k+n)+} \triangleq u_{k+n:k+N-1}$. If these bounds do not overlap (see Fig. \ref{fig: bounds no overlap}), one can guarantee to select the optimal action sequence as in \eqref{eq: bellman objective approx}.

Our general Nonmyopic Distilled Data Association BSP (ND2A-BSP) approach is presented in Algorithm \ref{alg: nonmyopic distliied data association BSP}. The algorithm receives a belief tree \skeleton; a heuristic function $h$ used to select the subsets of hypotheses in each belief tree node, i.e. defines $b_{k+n}^s$; and a decision rule $R$ which decides whether the considered subsets are enough, e.g. when no overlap between bounds is required or when calculations exceed a user defined time threshold, providing anytime performance guarantees. The algorithm returns the best action sequence, given the computational constraints, and an upper bound on the loss in solution quality.

It is worth mentioning that our approach can be adapted to a setting where the belief tree construction is coupled with Q function estimates, e.g. using MCTS and Upper Confidence Bound (UCB) techniques \cite{Silver10nips}, following a similar approach to the one presented in \cite{Sztyglic21arxiv_b}. However, we emphasize that as the belief tree \skeleton approximates \eqref{eq:general objective recursive} via samples, our method provides performance guarantees with respect to that specific \skeleton, i.e. with respect to \eqref{eq: bellman objective approx}. Not to be confused with the asymptotic guarantees of MCTS approaches, with respect  to the theoretical problem \eqref{eq:general objective recursive}, which is an entirely different aspect not related to the approach presented in this paper.
\vspace{-14pt}
\begin{algorithm}
	\smaller
	\SetAlgoLined	
	\KwIn{belief tree \skeleton $T$, simplification heuristic $h$, decision rule $R$}
	\KwOut{action sequence $u^*$, loss}
	\SetKwFunction{FMain}{ND2A-BSP}
	\SetKwProg{Fn}{Function}{:}{}
	\Fn{\FMain{$T,h, R$}}{
		$LB^*,UB^*, loss = \mathtt{PLAN} \left( T.root,h,R \right)$ \\
		$u^* \leftarrow$ corresponding to $LB^*,UB^*$ \\
		\KwRet $u^*, loss$
	}
	\SetKwFunction{FPlan}{PLAN}
	\SetKwProg{Fn}{Function}{:}{}
	\Fn{\FPlan{$Node,h,R$}}{
		$Node.b_{k+n}^s \leftarrow h \left( Node \right)$ \\
		\If{Node \normalfont{is a leaf}}{
			\KwRet $\underline{c} \left( Node.b_{k+n}^s \right) , \bar{c} \left( Node.b_{k+n}^s \right), 0$ \tcp*[f]{$loss=0$ at leaf} \\
		}
	    $Node.bounds = \emptyset$ \\ 
		\ForEach{\normalfont{child} $C$ \normalfont{of} $Node$}{
			$lb,ub,loss \leftarrow$ ND2A-BSP($C,h,R$) \\
			$LB \leftarrow \underline{c} \left( Node.b_{k+n}^s \right) + lb $\tcp*[f]{objective lower bound \eqref{eq: recrusive objective bounds}} \\
			$UB \leftarrow \bar{c} \left( Node.b_{k+n}^s \right) + ub $\tcp*[f]{objective upper bound \eqref{eq: recrusive objective bounds}} \\  
			$Node.bounds = Node.bounds \cup \left(LB,UB \right)$ \\
		}
		\While{$R \left( Node.bounds \right)$ \normalfont{is not satisified}} {
			ND2A-BSP($Node,h,R$)  \tcp*[f]{further simplification is needed}}
		$LB^*,UB^*, loss \leftarrow Node.bounds$ \\
		\KwRet $LB^*,UB^*, loss$
	}
	\caption{\smaller Generic Nonmyopic Distilled Data Association BSP}
	\label{alg: nonmyopic distliied data association BSP}
	\footnotetext{test}
\end{algorithm}
\vspace{-14pt}

We now analyze different settings, within inference and planning, where the agent either has or does not have hard budget constraints. To the best of our knowledge, this is the first time that these aspects are addressed in works that attempt to reduce the computational complexity of the planning problem. The differences between the considered settings are summarized in Table \ref{table: cases budget table}. 

\begin{table} [!h]
	\tiny
	\centering
	\begin{tabular} {|c|c|c|}
		\hline
		& budget constraints in inference & budget constraints in planning \\
		\hline 
		Case 1 & \XSolidBrush &  \XSolidBrush \\ 
		Case 2 & \XSolidBrush & \Checkmark  \\
		Case 3 & \Checkmark &   \XSolidBrush \\
		Case 4 & \Checkmark & \Checkmark  \\		
		\hline
	\end{tabular}
	\caption{A summary of the considered scenarios, with respect to budget constraints on the number of supported hypotheses in each algorithm, for each considered case. Cases 1\&2 are presented in Section \ref{section: no budget constraints in inference} while cases 3\&4 are presented in section \ref{section: hard budget constraint in iference}}.
	\label{table: cases budget table}
	\vspace{-22pt}
\end{table}

\subsection{No budget constraints in inference} \label{section: no budget constraints in inference}
\begin{figure*} [h]
	\begin{subfigure}{0.5\textwidth}
		\centering
		\includegraphics[scale=0.25]{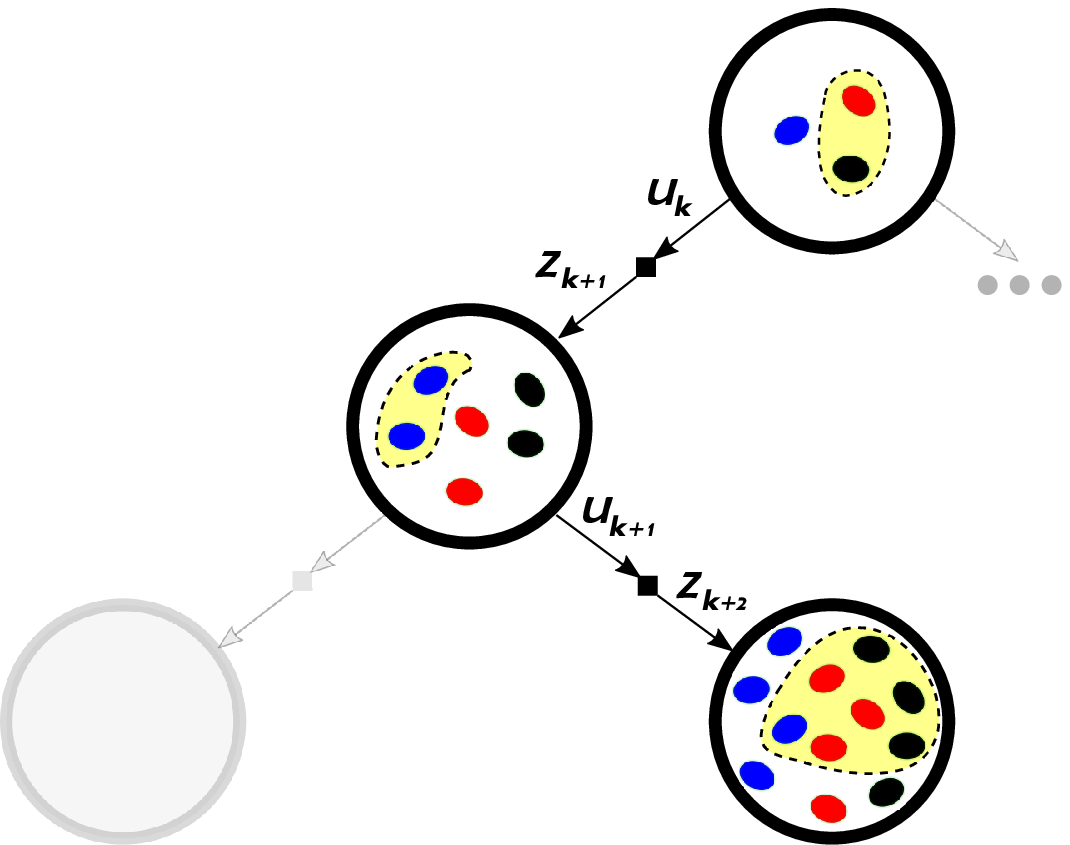}
		\caption{}
		\label{fig: no budget in inference nor in planning}
	\end{subfigure}
	\begin{subfigure}{0.5\textwidth}
		\centering
		\includegraphics[scale=0.25]{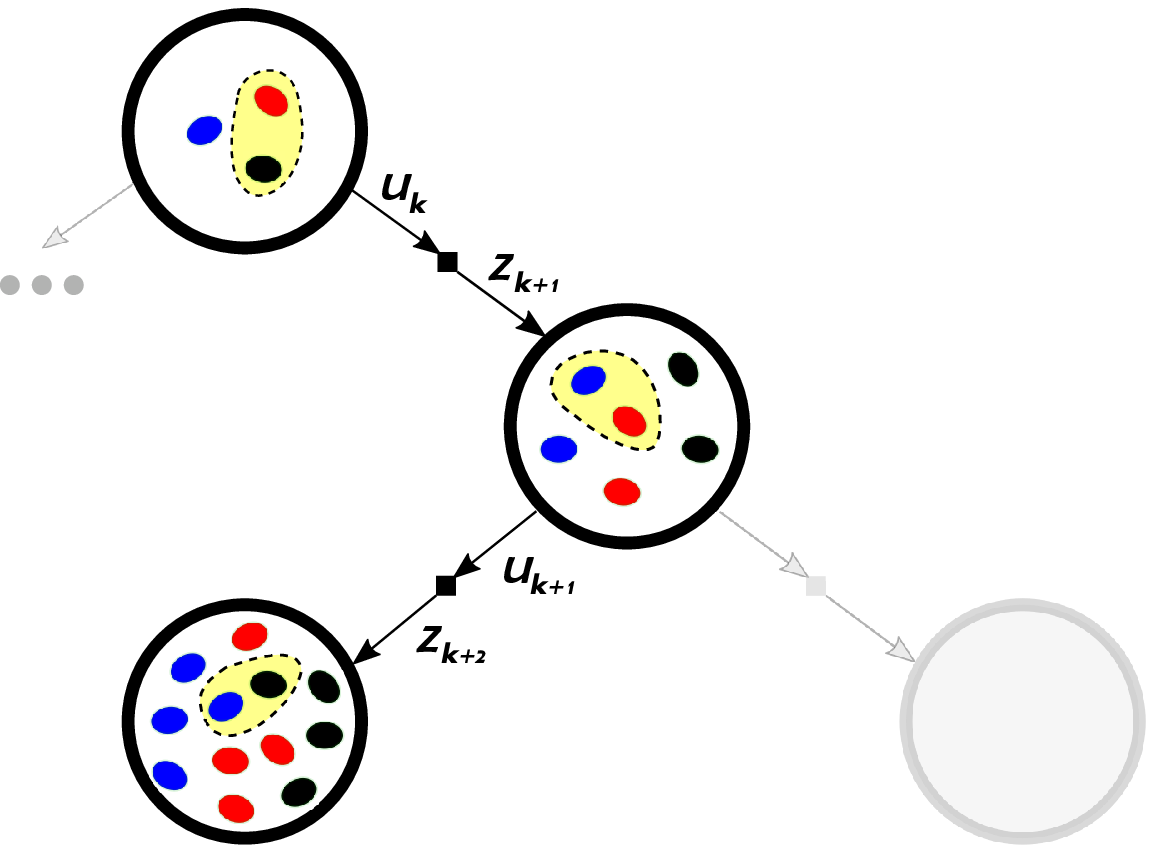}
		\caption{}
		\label{fig: no budget in inference, budget in planning}
	\end{subfigure}
	\caption{No budget constraints in inference. Colors denote components generated from previous time steps. (a) Planning without budget constraints, the algorithm can choose any subset of components, highlighted in yellow, in each node to evaluate the bounds; (b) With budget constraints in planning, each subset selection is bounded in size by \budget $=2$.}
	\label{fig: no budget constraints in inference}
	\vspace{-16pt}
\end{figure*}

In this section we assume that there are no constraints in inference, i.e. each belief tree node can theoretically hold every possible hypothesis within the planning horizon. The objective of inference however is different than the main goal of BSP. In inference the agent tries to represent the considered state as accurately as possible while in planning the goal is to retrieve the optimal action sequence or policy. As such, in this setting, the problems are decoupled (see Fig. \ref{fig: no budget in inference nor in planning}).

We now further separate between two cases, when the planning algorithm either has budget constraints or not. In both cases, each belief tree node still has an exponential number of components, which we avoid calculating explicitly.

\subsubsection{Case 1} 
With no budget constraints in planning we propose bounding the cost function as 
\begin{equation} \label{eq: cost bounds theoretical}
	\underline{c} \left( b_k, u_{k+}, Z_{(k+1)+}, b_{k+n}^s  \right) \leq c \left( b_{k+n}, u_{k+n} \right) \leq \bar{c} \left( b_k, u_{k+}, Z_{(k+1)+}, b_{k+n}^s \right).
\end{equation} 
where $u_{k+} \triangleq u_{k:k+n-1}$ and $Z_{(k+1)+} \triangleq Z_{k+1:k+n}$. A key difference from the approach presented in \cite{Sztyglic21arxiv} is that these bounds are not functions of $b_{k+n}$.

As the number of belief components grows exponentially we avoid calculating  $c \left( b_{k+n} \right)$. Instead, we calculate a \simplified \! \! belief $b_{k+n}^s$, using Bayesian updates via $u_{k:k+n-1}$ and $Z_{k+1:k+n}$, only for specific components from the prior belief $b_k$. This extends our proposed approach in \cite{Shienman22icra} to the nonmyopic case. Each \simplified \! \! belief is formally defined, using $M_{k+n}^s \subseteq M_{k+n}$ components, as
\begin{equation}
	b_{k+n}^s \triangleq \sum_{r \in M^s_{k+n}} w_{k+n}^{s,r} b_{k+n}^r \quad , \quad w_{k+n}^{s,r} \triangleq \frac{w_{k+n}^{r}}{w_{k+n}^{m,s}},
\end{equation} 
where \small $w_{k+n}^{m,s} \triangleq \sum_{m \in M^s_{k+n}} w^m_{k+n}$ \normalsize is used to re-normalize each corresponding weight. Most importantly, a \simplified \!\, belief $b_{k+n}^s$ is calculated using only a subset of hypotheses, i.e. without  calculating the posterior belief $b_{k+n}$.

Using Algorithm \ref{alg: nonmyopic distliied data association BSP} given a decision rule $R$, with no overlap between bounds \eqref{eq: recrusive objective bounds}, and a heuristic $h$, e.g. which chooses hypotheses greedily based on prior weights, we guarantee the selection of the optimal actions sequence, with respect to the specific belief tree, while reducing the computational complexity.

\subsubsection{Case 2} 
Under budget constraints in planning, the algorithm can use up to \budget  components, in each \simplified \! belief $b_{k+n}^s$, to calculate the bounds in \eqref{eq: cost bounds theoretical}. Yet, each subset of components is chosen independently w.r.t.~$b_{k+n}$ which develops exponentially, i.e.~hypotheses chosen in time steps $k+n$ and $k+n+i$ are not necessarily related (see Fig. \ref{fig: no budget in inference, budget in planning}).

In this setting, the number of possible distilled subsets for each $b_{k+n}^s$ is $\left| M_{k+n} \right| \choose \mathcal{C}$ which can be very high. Moreover, there are no guarantees that the bounds between candidate actions would not overlap. However, using the bounds in \eqref{eq: cost bounds theoretical}, our proposed approach can yield the worst-case loss in solution quality, i.e. provide performance guarantees (see Fig. \ref{fig: bounds overlap}). %In this work we consider a normalized loss, i.e. the upper bound over the loss in (see Fig. \ref{fig: bounds overlap}) is normalized by the interval defined by \eqref{eq: recrusive objective bounds} over \small $\hat{u}_{k:k+N-1}^*$. \normalsize 

\subsection{Hard budget constraints in inference} \label{section: hard budget constraint in iference}
In the previous section we only considered that the belief at the root of the tree is provided from inference. As the posterior beliefs within the constructed belief tree were with an exponentially increasing number of components, i.e. without budget constraints, the key idea was to avoid making explicit inferences. Instead, we calculated bounds that utilized, under budget constraints in planning, a fixed number of components. In practice, however, real world autonomous systems do not work that way. Instead, they are often required to operate in real time using inexpensive hardware with hard computational budget constraints in both inference and planning. 

Under hard budget constraints on the number of considered hypotheses in inference, the posterior belief in each belief tree node is determined by \eqref{eq: inference psi}, i.e. $| M_{k+n}^\psi | \leq $ \budget \! under some heuristic $h^{inf}$. Moreover, once a hypothesis is discarded in time step $k$ it is no longer considered in future time steps. Yet, the decision regarding which components to choose, while calculating the bounds in planning, depends on either if the heuristic in \eqref{eq: inference psi} is given or determined within planning. To the best of our knowledge, the latter is a novel concept never considered.

\begin{figure*} [h]
	\begin{subfigure}{0.5\textwidth}
		\centering
		\includegraphics[scale=0.25]{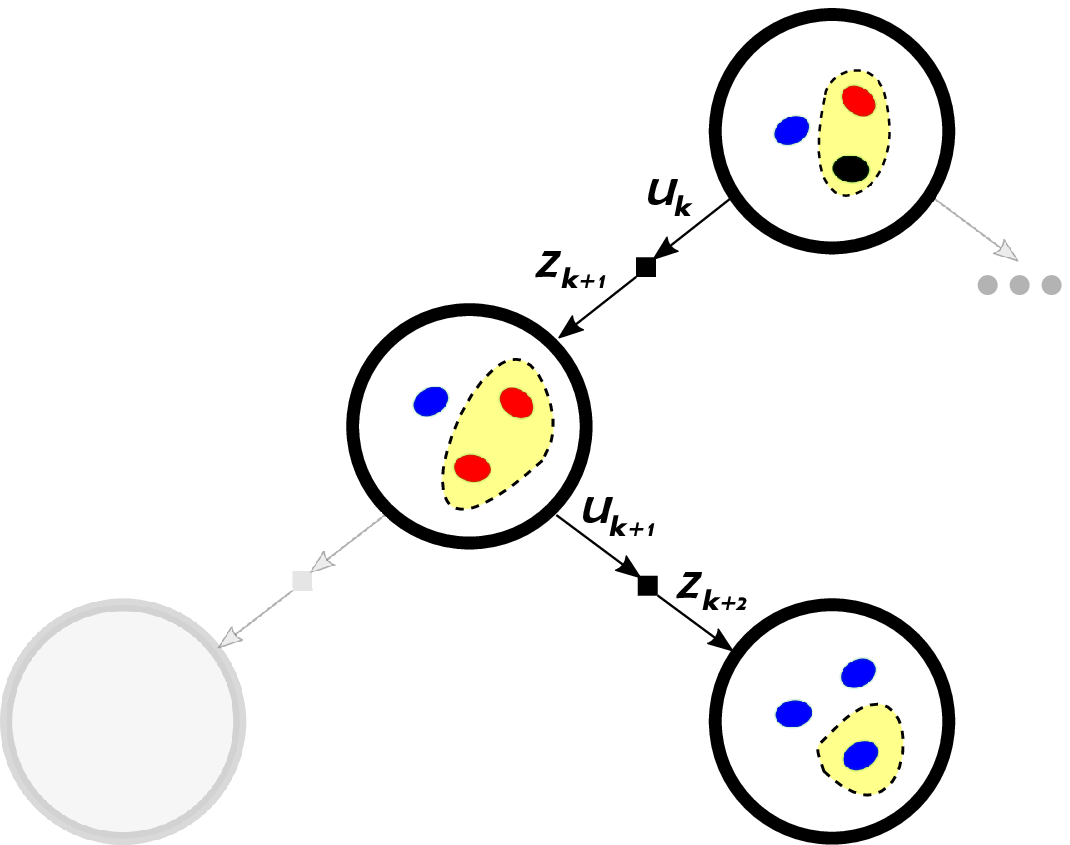}
		\caption{}
		\label{fig: planning given heuristict}
	\end{subfigure}
	\begin{subfigure}{0.5\textwidth}
		\centering
		\includegraphics[scale=0.25]{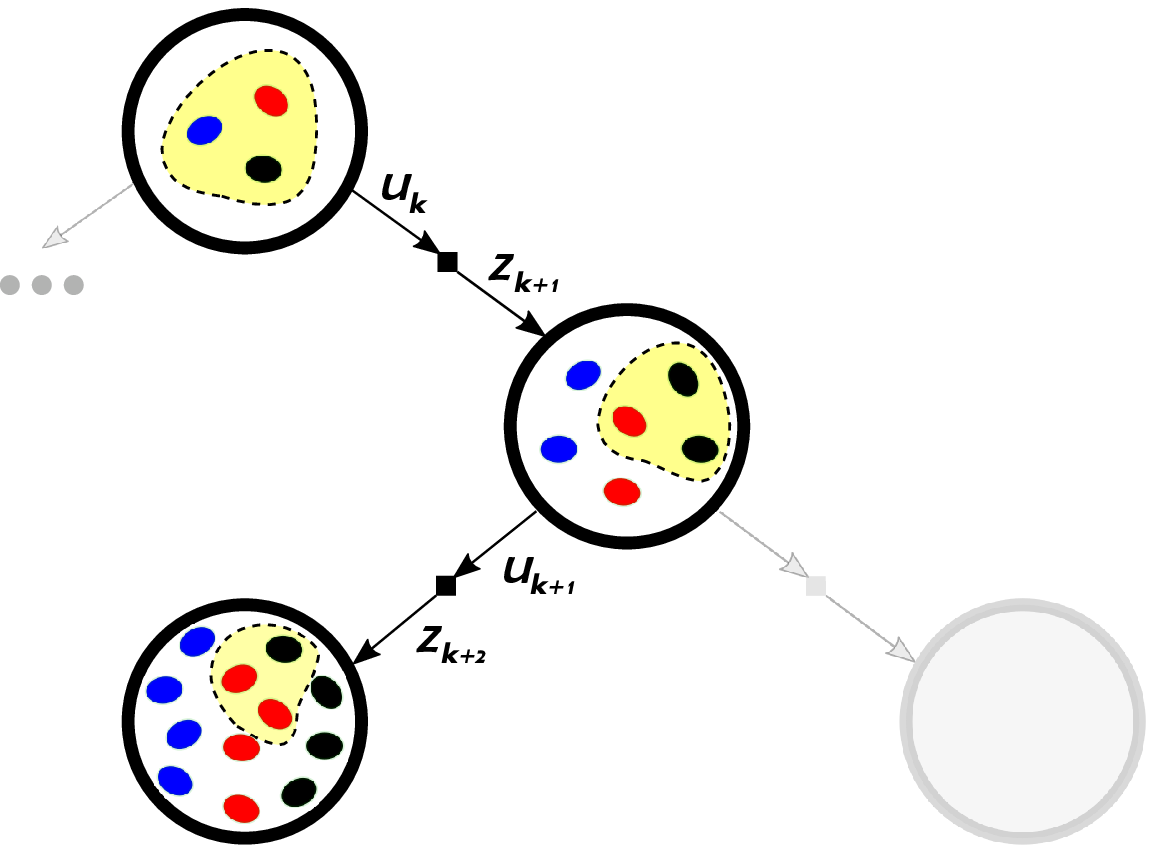}
		\caption{}
		\label{fig: planning decides heuristict}
	\end{subfigure}
	\caption{Hard budget constraints in inference. Colors denote components generated from previous time steps. (a) Planning given the heuristic in inference, the algorithm can only evaluate the bounds using components that represent how the belief would evolve in inference; (b) The planning algorithm is free to choose components under any valid heuristics in inference given the budget \budget. Each selected component in time step $k+n+1$ must originate from a selected component in time step $k+n$.}
	\label{fig: hard budget constraints in inference}
	\vspace{-16pt}
\end{figure*}

\subsubsection{Case 3} \label{sec:Planning under a given heuristics} 
In this setting we consider the heuristic in \eqref{eq: inference psi} to be given within planning, i.e. posterior belief tree nodes exactly represent how the belief would evolve in inference under \eqref{eq: inference psi}. In contrast to Section \ref{section: no budget constraints in inference}, as the number of components does not grow exponentially, we sample future observations according to \eqref{eq: sample from posterior} and construct the belief tree explicitly, i.e. perform inference in each node. 
Therefore, the planning algorithm can no longer choose any subset of components for each $b_{k+n}^s$, i.e. hypotheses discarded in time step $k+n$ cannot be considered in time step $k+n+1$ (see Fig. \ref{fig: hard budget constraints in inference}).
	
The bounds over the considered cost are now a function of the belief in the previous time step under \eqref{eq: inference psi}. Specifically, we rewrite them as
\small
\begin{equation} \label{eq: cost bounds under psi}
		\underline{c} \! \left( b_{k+n-1}^\psi, u_{k+n-1}, Z_{k+1},  b_{k+n}^s \right) \!  \! \leq \!  c \! \left( \! b_{k+n}^\psi, u_{k+n} \! \right) \!\!  \leq \! \bar{c} \left( \! b_{k+n-1}^\psi, u_{k+n-1}, Z_{k+1}, b_{k+n}^s  \! \right).
\end{equation} 
\normalsize
These bounds represent a recursive setting in contrast to the bounds in \eqref{eq: cost bounds theoretical}.

Using our approach iteratively in each time step, reduces the computational complexity of the considered cost function in planning while providing performance guarantees. As each posterior belief is determined by inference (Fig \ref{fig: planning given heuristict}), performance guarantees are with respect to the given heuristic in inference \eqref{eq: inference psi}.

\subsubsection{Case 4}
We now relax the assumption that the planning algorithm is confined to the specific heuristic in \eqref{eq: inference psi}. Unlike in Case 2, where each subset of components can be used in each node to calculate the bounds, this setting has an additional constraint. We formulate this by representing the bounds from \eqref{eq: cost bounds theoretical} in two consecutive time steps
\begin{equation} \label{eq: case 4 bounds}
	\begin{split}
	&\underline{c} \left( b_k, u_{k+}, Z_{(k+1)+}, b_{k+n}^s  \right) \leq c \left( b_{k+n} \right) \leq \bar{c} \left( b_k, u_{k+}, Z_{(k+1)+}, b_{k+n}^s \right), \\ 
	&\underline{c} \left( b_k, u_{k+}, Z_{(k+1)+}, b_{k+n+1}^s  \right) \leq c \left( b_{k+n+1} \right) \leq \bar{c} \left( b_k, u_{k+}, Z_{(k+1)+}, b_{k+n+1}^s \right), \\
	&\text{s.t.} \left| M_{k+n}^s \right| , \left| M_{k+n+1}^s \right| \leq  \mathcal{C} \quad \text{and} \quad  \forall b_{k+n+1}^{s,ij} \in b_{k+n+1}^s \Rightarrow b_{k+n}^{s,j} \in b_{k+n}^s,
\end{split}
\end{equation}
where $b_{k+n}^{s,j}$ denotes the $j$th hypothesis in the \simplified \!\, subset $b_{k+n}^s$ and  $b_{k+n+1}^{s,ij}$ denotes the $i$th hypothesis in the \simplified \!\, subset $b_{k+n+1}^s$, originated from $b_{k+n}^{s,j}$, i.e. as in Fig. \ref{fig:exponential growth}. 

The components chosen in the sequence of bounds \eqref{eq: case 4 bounds} which minimizes the loss, w.r.t.~the original problem, define a heuristic $h^{p\star}$ (see Fig. \ref{fig: results scenario4}), which is valid in inference.
The heuristic $h^{p\star}$ can be used with any BSP approach to solve \eqref{eq: bellman objective approx} and to reduce computational complexity, using our approach, as described in Case 3.
To the best of our knowledge, leveraging $h^{p\star}$ is a novel concept. We note that while $h^{p\star}$ minimizes the loss in planning, it is generally different than $h^{inf}$. As such, the implications of utilizing such heuristic in inference are not straightforward. The study of such mechanism is left for future research.

\subsection{The cost function}
While the formulation thus far was for a general cost function, in this section we focus on active disambiguation of hypotheses. Specifically, we utilize the Shannon entropy, defined over posterior belief components weights. The cost for
a belief $b_{k+n}$ with $M_{k+n}$ components is thus given by $\mathcal{H}_{k+n} \triangleq c\left( b_{k+n} \right) = - \sum_{r \in M_{k+n}} \frac{ w^{r}_{k+n}}{\eta_{k+n}} log \left(\frac{ w^{r}_{k+n}}{\eta_{k+n}} \right)$, where $\eta_{k+n} \triangleq \sum_{r \in M_{k+n}} w_{k+n}^r$. Similarly, for a simplified belief $b_{k+n}^s$ with $M_{k+n}^s \subseteq M_{k+n}$ the cost is given by $\mathcal{H}_{k+n}^s \triangleq  c\left( b_{k+n}^s \right)  = - \sum_{r \in M_{k+n}^s}  w^{s,r}_{k+n} log \left( w^{s,r}_{k+n} \right)$.

To allow fluid reading, proofs for all theorems and corollaries are given in the supplementary material \cite{Shienman22isrr_Supplementary}.

\begin{theorem} \label{theorem: entropy relation}
	For each belief tree node representing a belief $b_{k+n}$ with $M_{k+n}$ components
	and a subset $M_{k+n}^s \subseteq M_{k+n}$ the 
	cost can be expressed by
	\begin{equation} \label{eq: entropy relation}
		\mathcal{H}_{k+n} = \frac{ w_{k+n}^{m,s} }{\eta_{k+n}} \left[ \mathcal{H}^s_{k+n} + log \left( \frac{\eta_{k+n}}{w_{k+n}^{m,s} } \right) \right] - \sum_{r \in \neg M_{k+n}^s} \frac{w_{k+n}^{r}} {\eta_{k+n}} log \left( \frac{w_{k+n}^{r}} {\eta_{k+n}} \right), 	
	\end{equation}
	where  $\neg M_{k+n}^s \triangleq M_{k+n} \setminus M_{k+n}^s$.
\end{theorem}
Using Theorem \ref{theorem: entropy relation}, we derive bounds for $\mathcal{H}_{k+n}$ which
are computationally more efficient to calculate as we only
consider a subset of hypotheses. However, as evaluating $\eta_{k+n}$ requires by definition evaluating all
posterior components weights, which we do not have access to, we need to bound this term as well (denoted below as $\mathcal{LB}\left[{\eta_{k+n}}\right]$ and $\mathcal{UB}\left[{\eta_{k+n}}\right]$).

\begin{theorem} \label{theorem:entropy bounds}
	Given a subset of components $M_{k+n}^s \subseteq M_{k+n}$, the cost term in each belief tree node is bounded by
	\begin{equation} \label{eq:entropy lower bound}
		\mathcal{LB} \left[ \mathcal{H}_{k+n} \right] = \frac{ w_{k+n}^{m,s} }{\mathcal{UB}\left[ \eta_{k+n} \right]} \left[ \mathcal{H}^s_{k+n} + log \left( \frac{\mathcal{LB}\left[ \eta_{k+n} \right]}{w_{k+n}^{m,s}} \right) \right],
	\end{equation}
	\begin{multline} \label{eq:entropy upper bound}
		\mathcal{UB} \left[ \mathcal{H}_{k+n} \right] = \frac{ w_{k+n}^{m,s} }{\mathcal{LB}\left[ \eta_{k+n} \right]} \left[ \mathcal{H}^s_{k+n} + log \left( \frac{\mathcal{UB}\left[ \eta_{k+n} \right]}{w_{k+n}^{m,s} } \right) \right] - \bar{\gamma} log \left(\frac{\bar{\gamma}}{\left| \neg M_{k+n} \right|}  \right),
	\end{multline}
    where $\bar{\gamma} = 1 - \sum_{r \in M_{k+n}^s} \frac{w_{k+n}^r}{\mathcal{UB}\left[\eta_{k+n}\right]}$ and $\left|\neg M_{k+n}^s \right|> 2$.
\end{theorem} 
Furthermore, considering different levels of simplifications, i.e. adding belief components to $M_{k+n}^s$, these bounds converge.
\begin{corollary}
	The bounds
	in Theorem \ref{theorem:entropy bounds} converge to $\mathcal{H}_{k+n}$
	when $M_{k+n}^s \rightarrow M_{k+n}$
	\begin{equation}
		\underset{M_{k+n}^s \rightarrow M_{k+n}}{\text{lim}} \mathcal{LB} \left[\mathcal{H}_{k+n}\right] = \mathcal{H}_{k+n} = \mathcal{UB} \left[\mathcal{H}_{k+n}\right].
	\end{equation}
\end{corollary}
A recursive update rule is given in Section C of the supplementary material \cite{Shienman22isrr_Supplementary}. 

\begin{theorem} \label{theorem:eta bounds}
	Given a subset of components $M_{k+n}^s \subseteq M_{k+n}$, the term $\eta_{k+n}$, in each belief tree node, is bounded by
	\small
	\begin{equation} \label{eq:eta bounds}
		\mathcal{LB} \left[ \eta_{k+n} \right] = w_{k+n}^{m,s} \leq \eta_{k+n} \leq
		w_{k+n}^{m,s} + \left(\frac{\left| M_{k+n} \right|}{\left| M_k \right|} - \sum_{r \in M_{k+n}^s} w_{k}^r \right) \prod_{i=1}^{n} \sigma^i = \mathcal{UB} \left[ \eta_{k+n} \right],
	\end{equation}
	\normalsize
	where $\sigma^i \triangleq \text{max} \left( \mathbb{P}\left( Z_{k+i}| x_{k+i} \right) \right)$ and $w_k^r$ is the prior weight at time $k$ for every component in $M_{k+n}^s$ at time $k+n$.
\end{theorem}
As in Theorem \ref{theorem:entropy bounds}, since we only consider a subset of hypotheses, these bounds are also computationally more efficient to calculate and  converge. We also note that specifically for Case 3, the bounds in Theorem \ref{theorem:entropy bounds} and Theorem \ref{theorem:eta bounds} are calculated iteratively in each time step $k+n$ given the belief $b_{k+n-1}^\psi$ as presented in \eqref{eq: cost bounds under psi}.

\begin{corollary}
	The bounds
	in Theorem \ref{theorem:eta bounds} converge to $\eta_{k+n}$
	when $M_{k+n}^s \rightarrow M_{k+n}$
	\begin{equation}
		\underset{M_{k+n}^s \rightarrow M_{k+n}}{\text{lim}} \mathcal{LB} \left[\eta_{k+n}\right] = \eta_{k+n} = \mathcal{UB} \left[\eta_{k+n}\right].
	\end{equation}
\end{corollary}
A recursive update rule is given in Section C of the supplementary material \cite{Shienman22isrr_Supplementary}.

% results
\section{Results}
We evaluate the performance of our approach for the different cases presented in Section \ref{sec: methods}. Our prototype implementation uses the
GTSAM library \cite{Dellaert12tr}.
\begin{figure*} [h]
	\begin{subfigure}[t]{0.3\linewidth}
		\centering
		\includegraphics[scale=0.16]{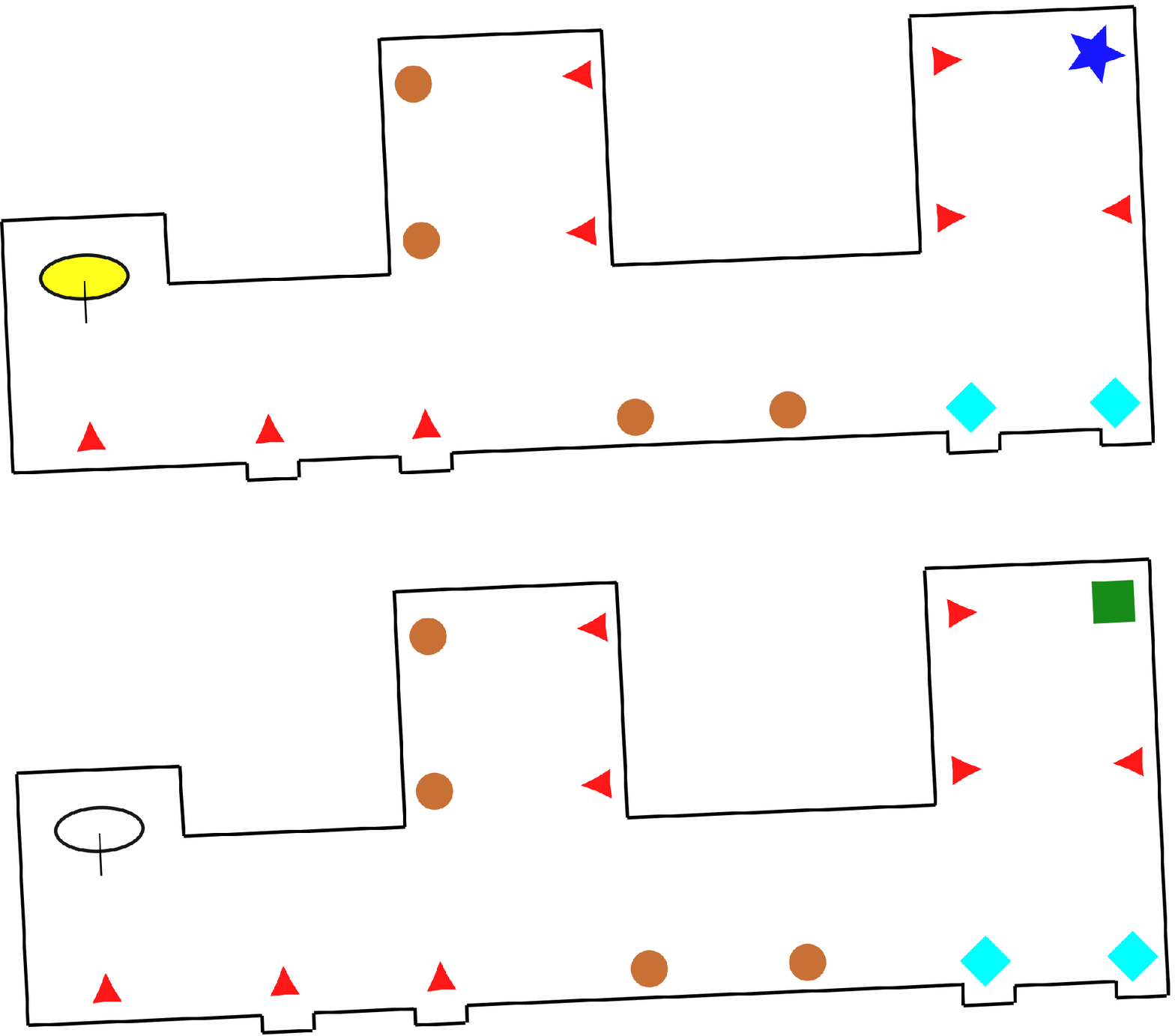}
		\caption{}
		\label{fig: floor plan}
	\end{subfigure}
	\hfill
	\begin{subfigure}[t]{0.45\linewidth}
		\centering
		\includegraphics[scale=0.25]{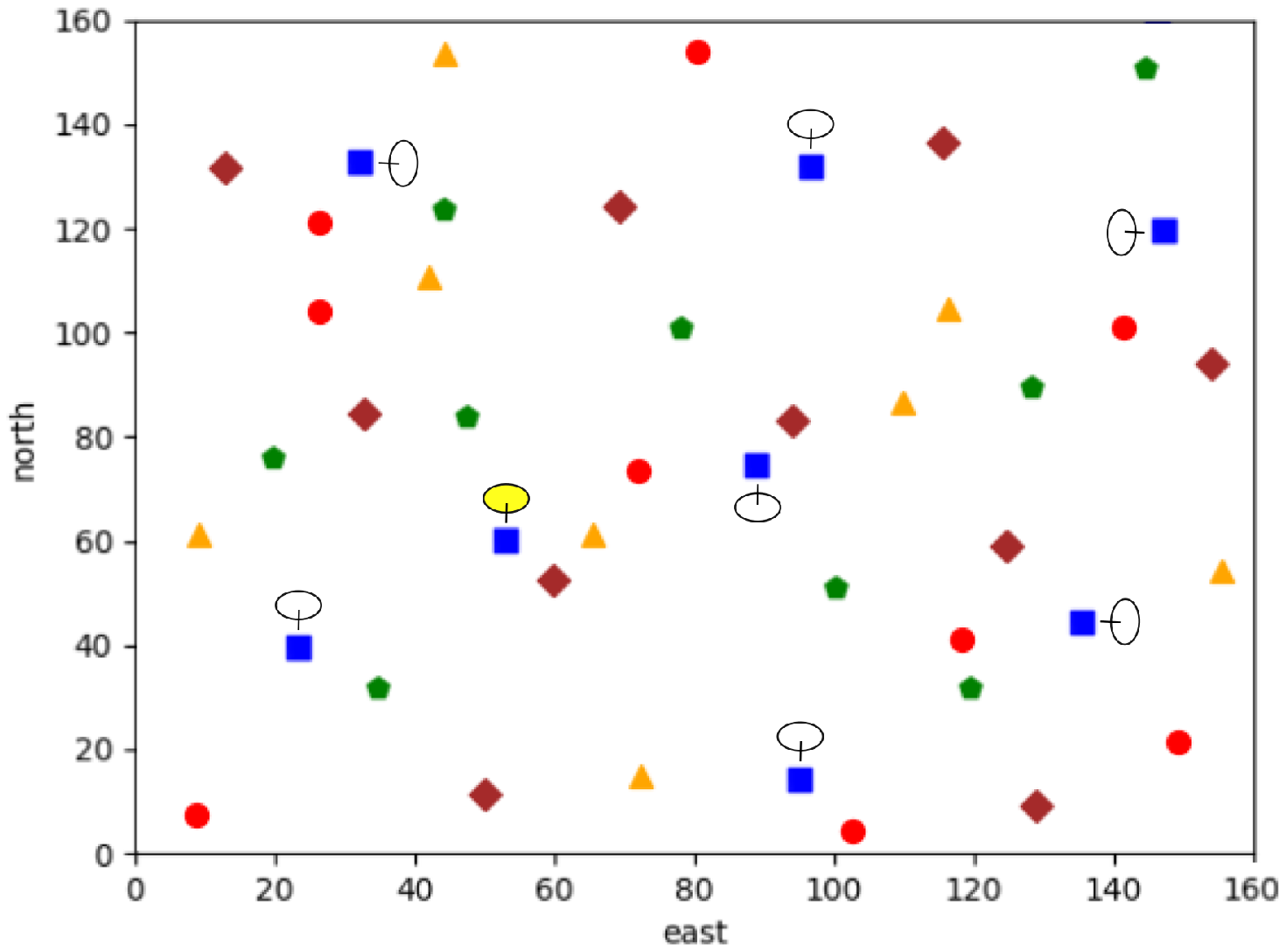}
		\caption{}
		\label{fig: random 2d}
	\end{subfigure}
	\hfill
	\begin{subfigure}[t]{0.20\linewidth}
		\centering
		\includegraphics[scale=0.18]{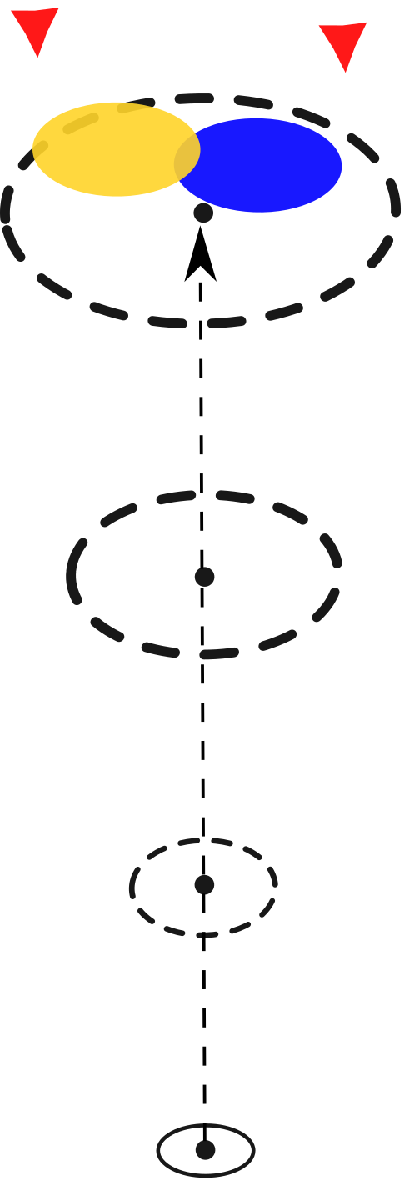}
		\caption{}
		\label{fig: da in planning}
	\end{subfigure}
	\caption{(a) The {\tt floors} environment where $F$ identical floors represent different prior hypotheses. Each floor contains a unique landmark. The true location of the agent is highlighted in yellow; (b) The {\tt 2d\_random} environment with many identical landmarks. The agent is initially placed in front of a blue square with no other prior information; (c) A planning session where ambiguous data association results in two hypotheses denoted by the yellow and blue ellipses.}
	\label{fig: simulation environment}
	\vspace{-16pt}
\end{figure*}
Our considered scenarios represent highly ambiguous environments containing perceptually identical landmarks in different locations.
In our first scenario, {\tt floors}, the agent is initially located in one of $F$ floors such that each floor contains a unique landmark, specific to that floor (Fig. \ref{fig: floor plan}). In our second scenario, {\tt 2d\_random}, the agent is initially placed in a random environment in front of a blue square (Fig. \ref{fig: random 2d}). Both scenarios can be considered as versions
of the kidnapped robot problem. With no other prior information, the initial belief, in both cases, is multi-modal containing $\left| M_0 \right|$ hypotheses. The agent captures the environment using range measurements containing a class identifier, e.g. red triangle or green square. When the agent receives a measurement to some landmark which is ambiguous, i.e. it can theoretically be generated from more than one landmark, the number of hypotheses grows (see Fig. \ref{fig: da in planning}). The number of identical landmarks can be adjusted to represent higher ambiguity, increasing the number of considered hypotheses. The agent's goal is to disambiguate between hypotheses by solving the corresponding BSP problem  \eqref{optimization objective} at each planning session using entropy over posterior belief components weights as a cost function.

\begin{figure*} [h]
	\begin{subfigure}{0.3\linewidth}
		\centering
		\includegraphics[scale=0.2]{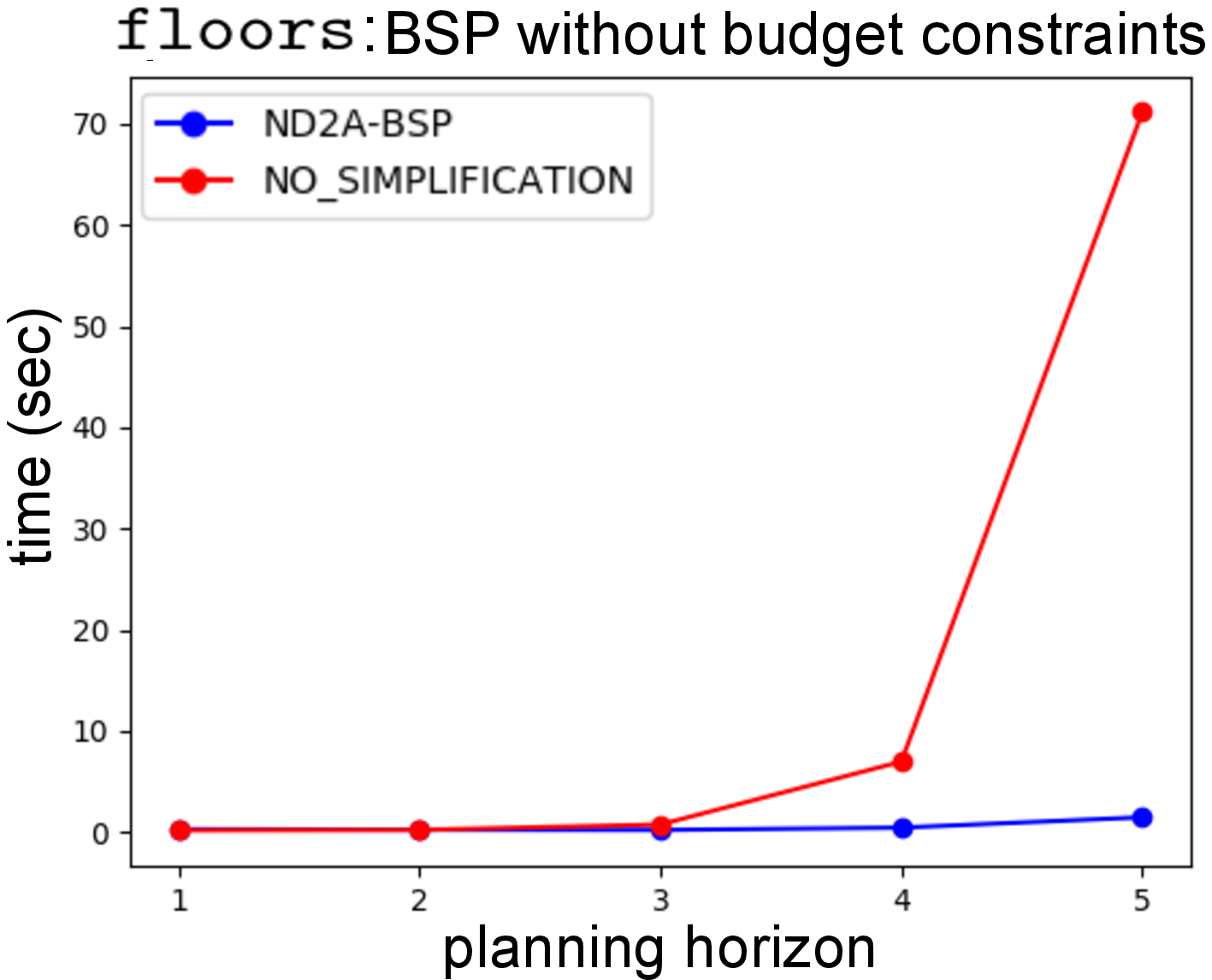}
		\caption{}
		\label{fig: results scenario1}
	\end{subfigure}
	\hfill
	\begin{subfigure}{0.3\linewidth}
		\centering
		\includegraphics[scale=0.2]{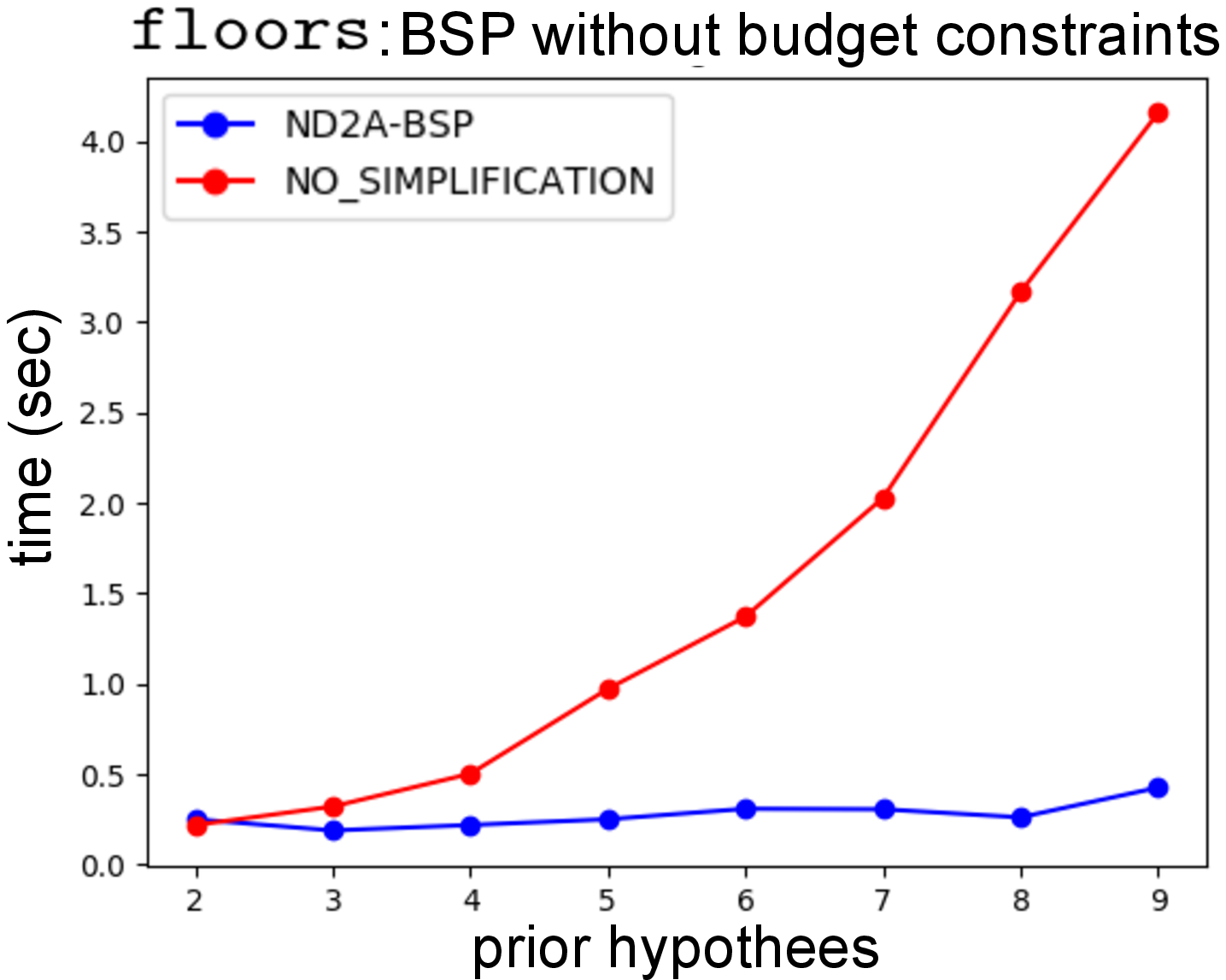}
		\caption{}
		\label{fig: results scenario1b}
	\end{subfigure}
	\hfill
	\begin{subfigure}{0.3\linewidth}
		\centering
		\includegraphics[scale=0.2]{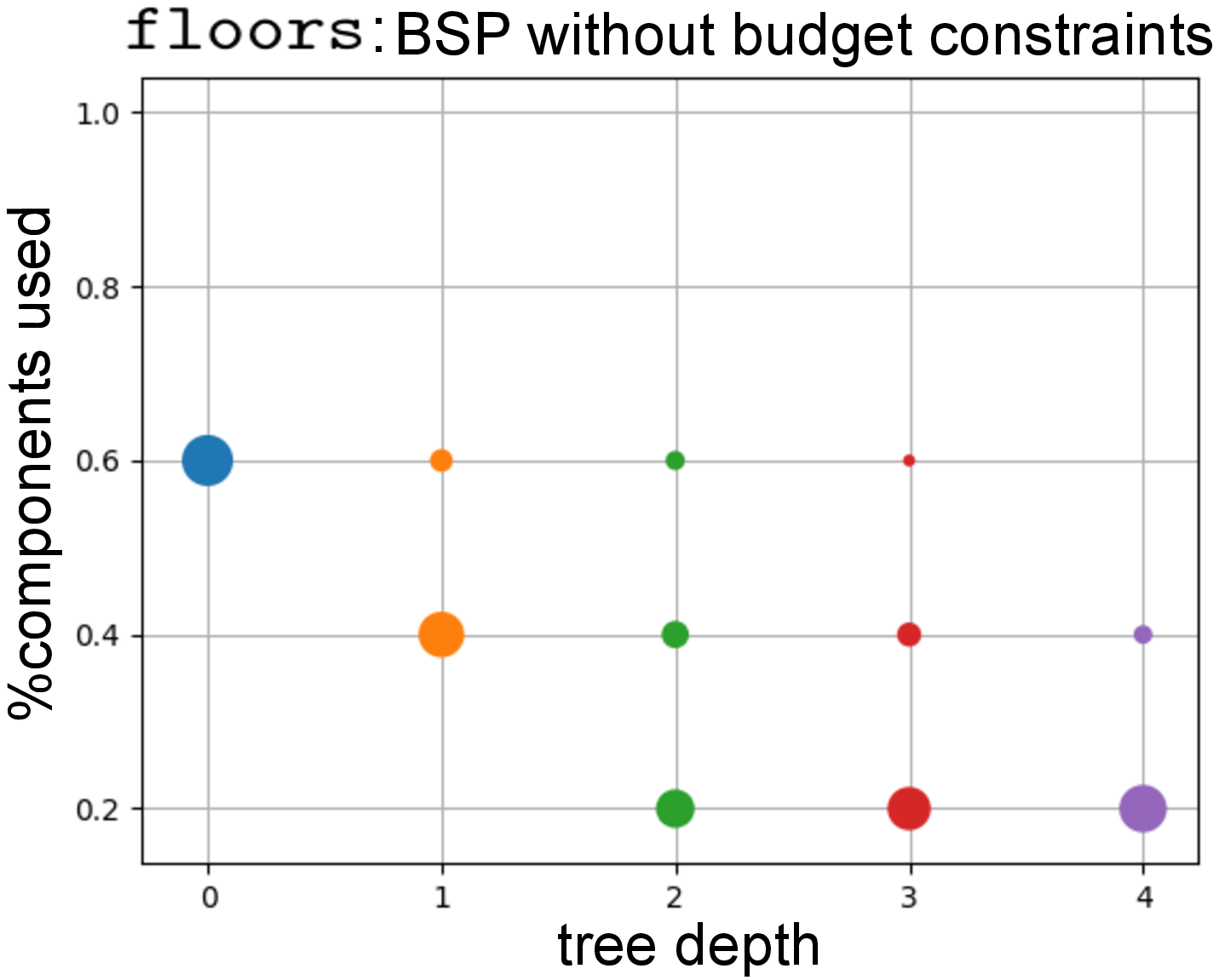}
		\caption{}
		\label{fig: results scenario1c}
	\end{subfigure}
	\begin{subfigure}{0.3\linewidth}
		\centering
		\includegraphics[scale=0.2]{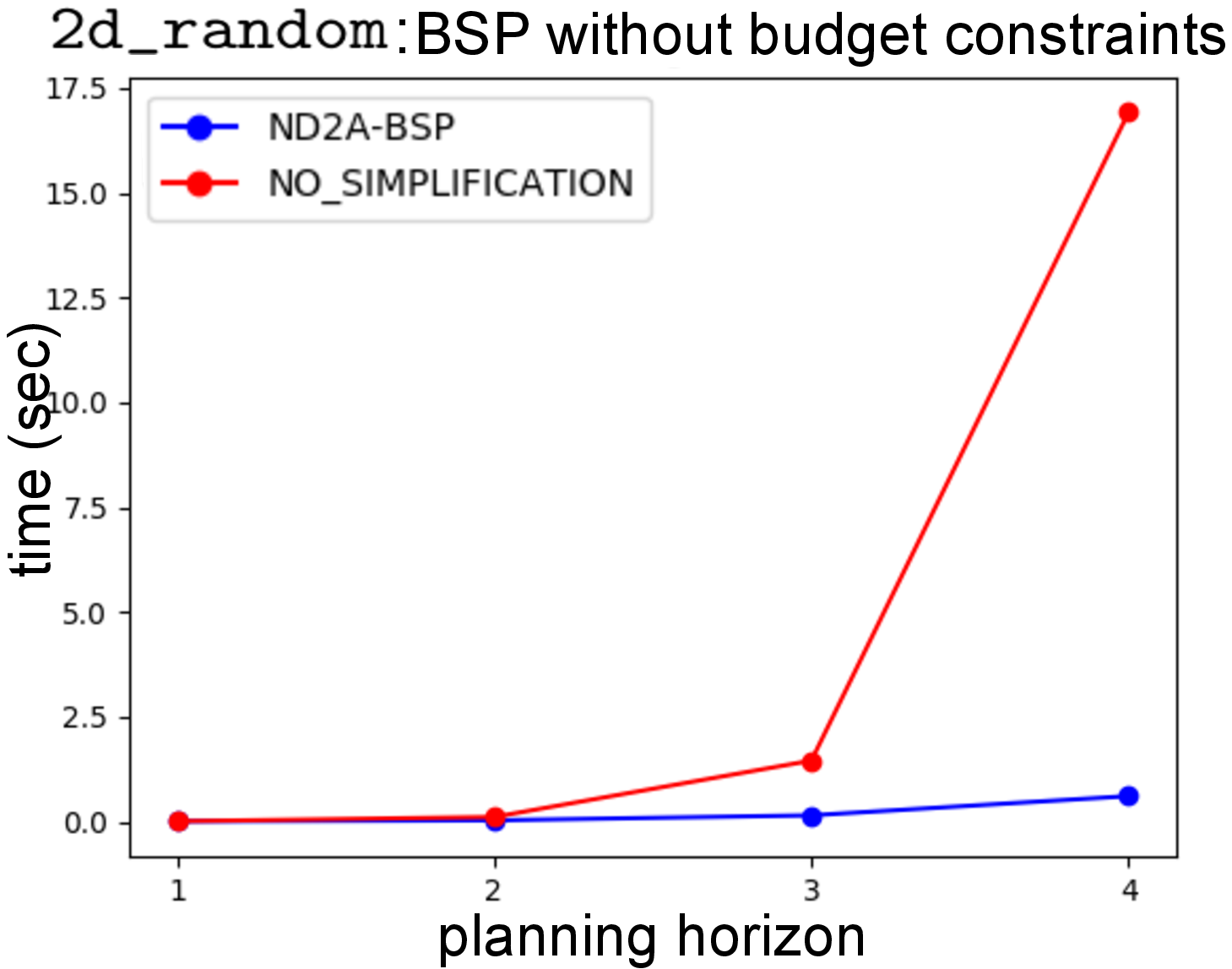}
		\caption{}
	\end{subfigure}
	\hfill
	\begin{subfigure}{0.3\linewidth}
		\centering
		\includegraphics[scale=0.2]{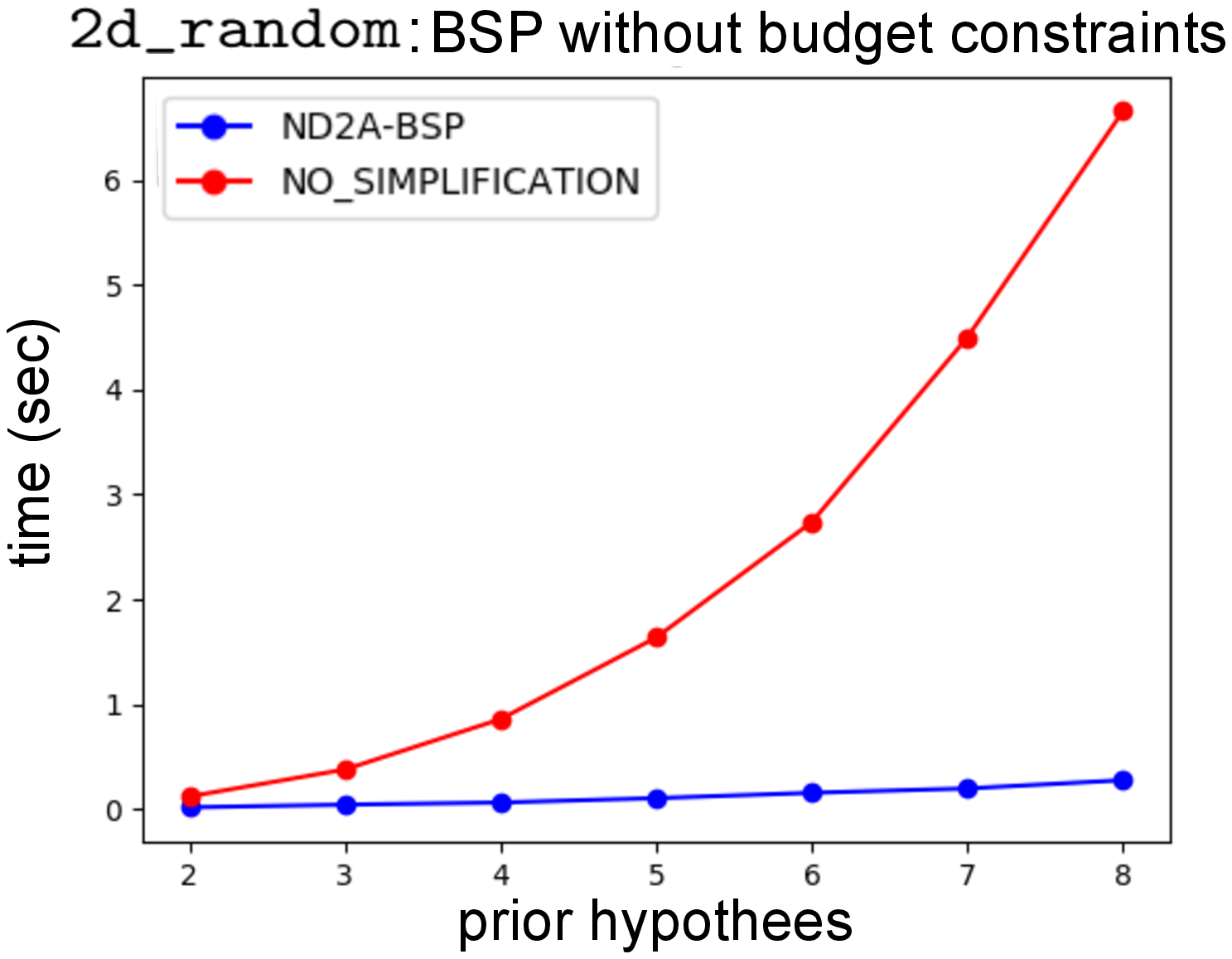}
		\caption{}
	\end{subfigure}
	\hfill
	\begin{subfigure}{0.3\linewidth}
		\centering
		\includegraphics[scale=0.2]{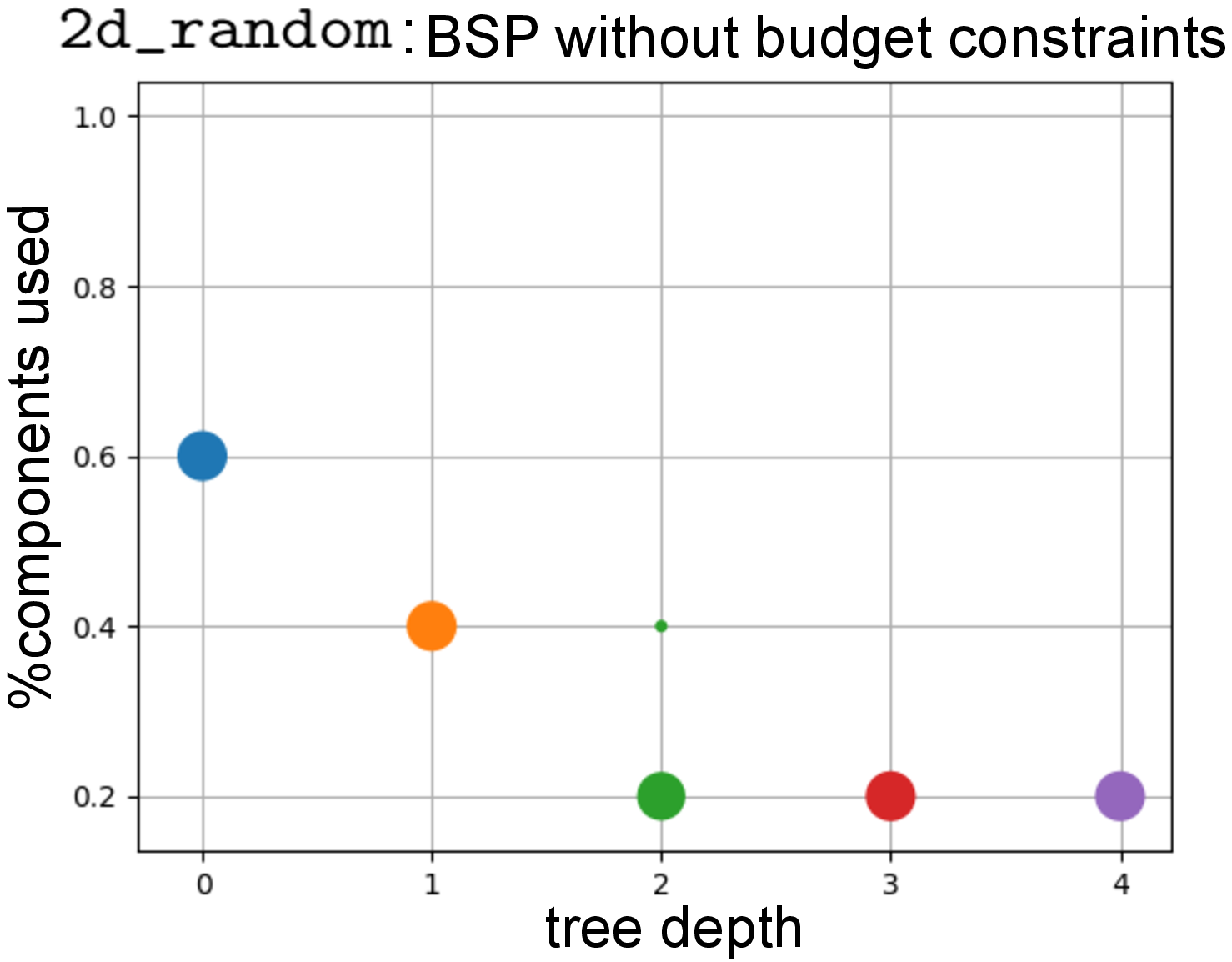}
		\caption{}
		\label{fig: results scenario1f}
	\end{subfigure}

	\caption{Case 1 study for {\tt floors} and {\tt 2d\_random} environments. All scenarios presented carry zero loss. (a),(d) Planning time as a function of the planning horizon. In both environments, all settings the considered 4 prior hypotheses; (b),(e) Planning time as a function of the number of prior hypotheses. In both environments, all settings considered a planning horizon of 3; (c),(f) \% components used to calculate bounds in each level of the belief tree. Circles scales are normalized as the number of nodes grows exponentially going down the tree.}
	\label{fig: main results without budgets}
	\vspace{-16pt}
\end{figure*}

In our first experiment we consider Case 1. We compare our approach with evaluating the cost function over the original belief, i.e. considering every possible future hypothesis. The heuristic in planning chooses the subset of components for each belief tree node greedily based on prior weights at time $k$. The decision rule $R$ was set as no overlap, i.e. no loss with guaranteed optimal solution. The computational merits of our approach are presented in Fig. \ref{fig: main results without budgets}. Moreover, in Fig. \ref{fig: results scenario1c},\ref{fig: results scenario1f} we can see that with a longer planning horizon the subset of hypotheses used for disambiguation becomes smaller. As more observations are utilized along the horizon, it is easier to discard wrong hypotheses in our considered cases.

\begin{figure*} [h]
	\begin{subfigure}{0.3\linewidth}
		\centering
		\includegraphics[scale=0.20]{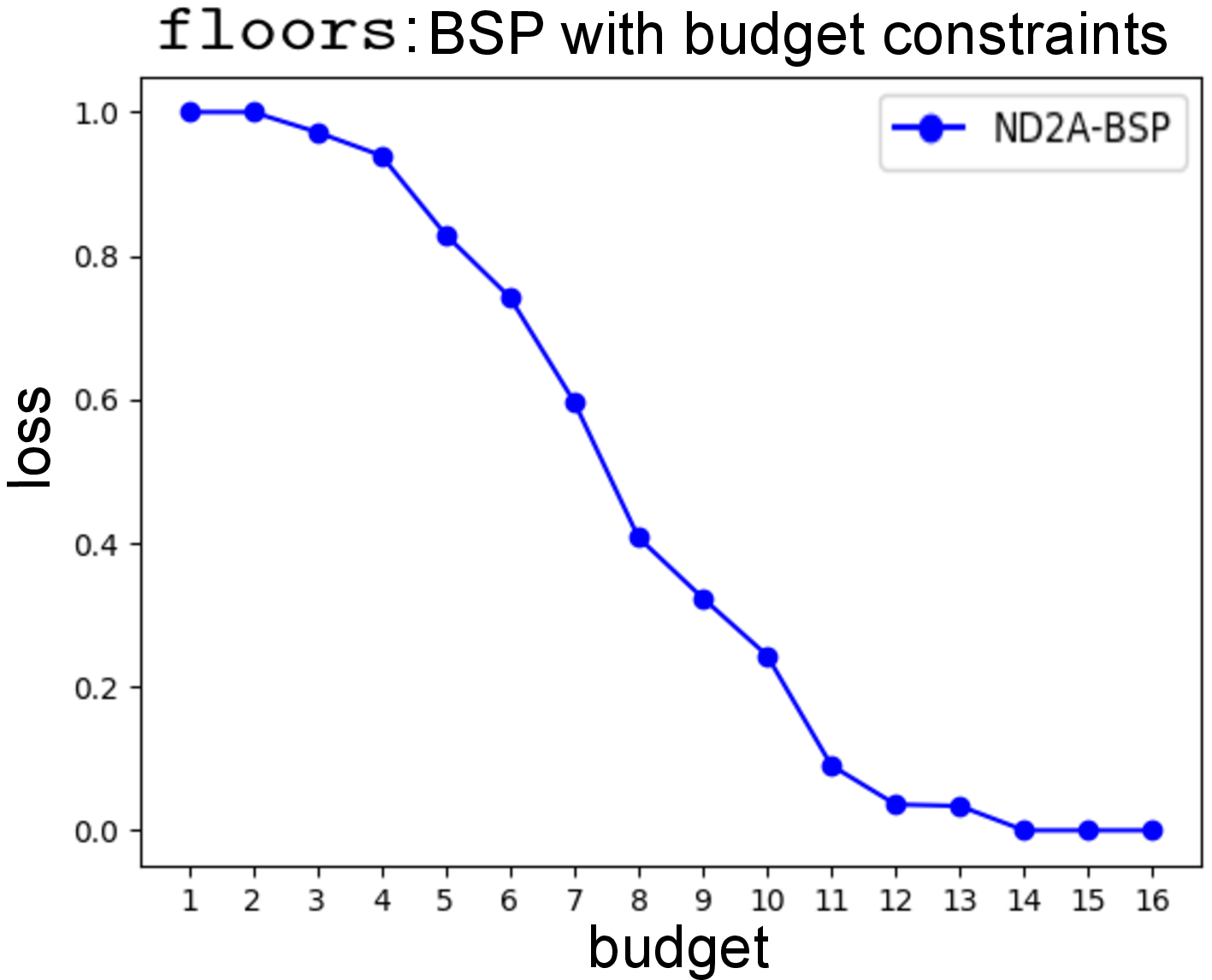}
		\caption{}
		\label{fig: results scenario2}
	\end{subfigure}
	\hfill
	\begin{subfigure}{0.3\linewidth}
		\centering
		\includegraphics[scale=0.20]{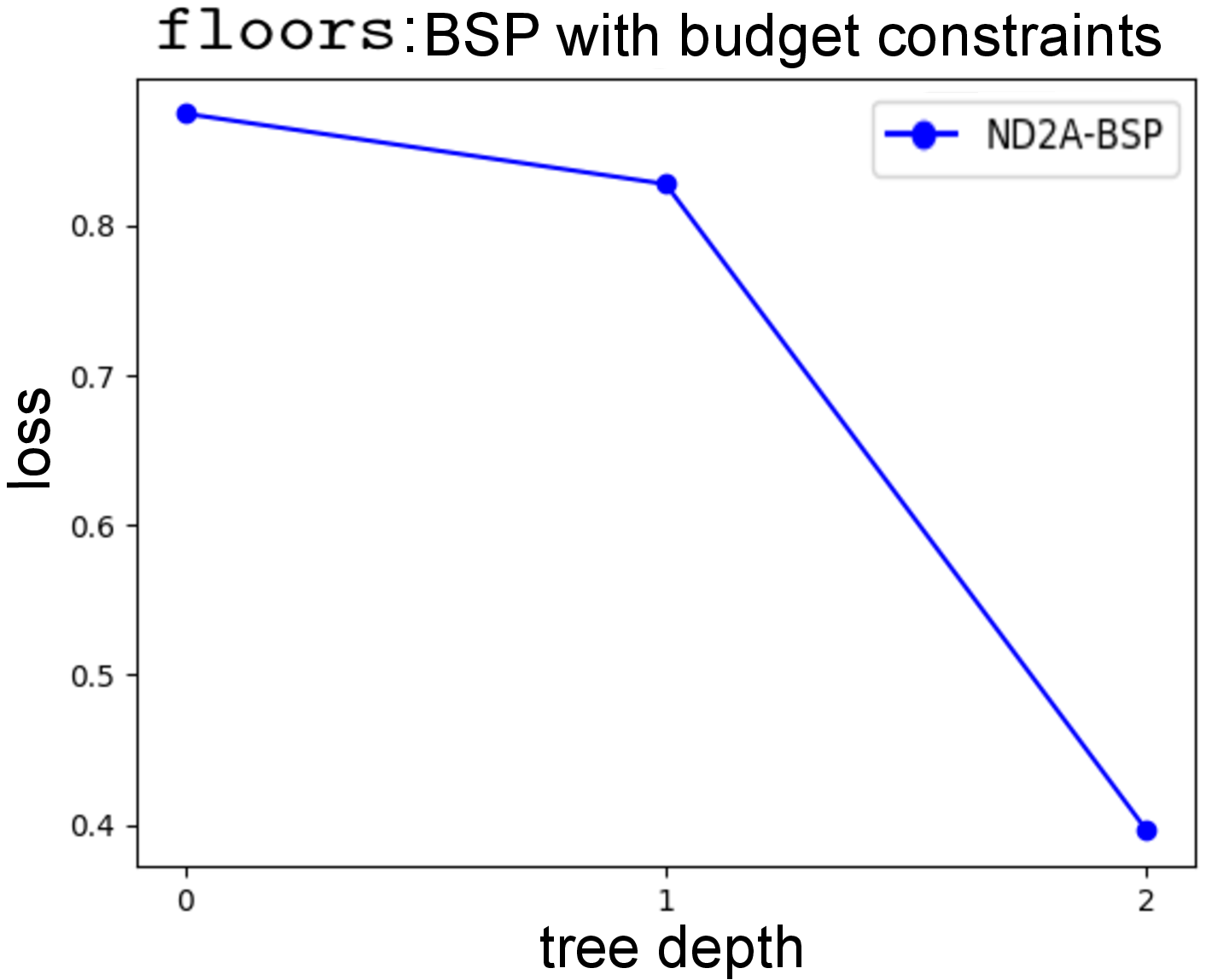}
		\caption{}
		\label{fig: results scenario2b}
	\end{subfigure}
	\hfill
	\begin{subfigure}{0.3\linewidth}
		\centering
		\includegraphics[scale=0.20]{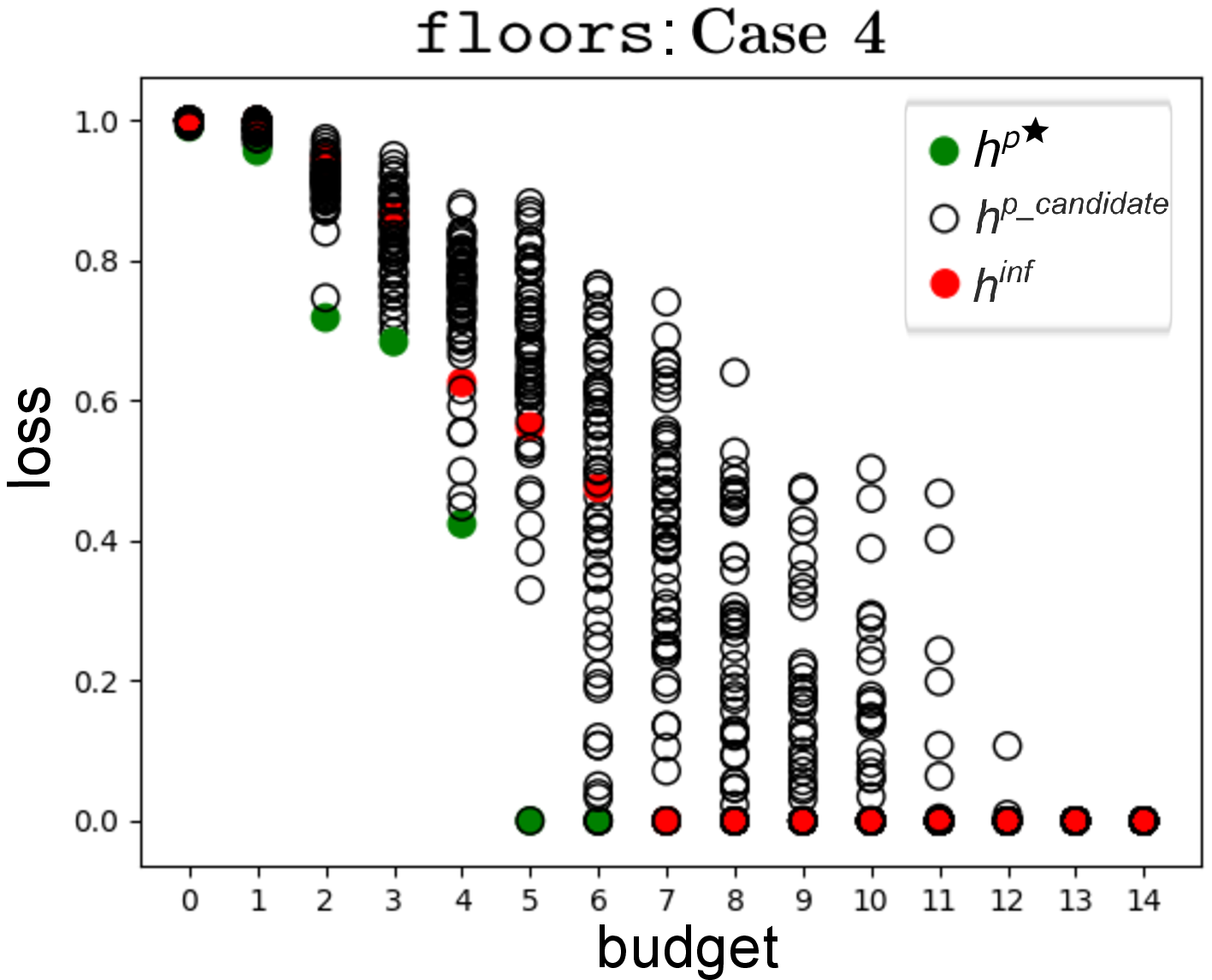}
		\caption{}
		\label{fig: results scenario4}
	\end{subfigure}
	\caption{(a) Normalized loss as a function of the size of the budget in Case 2. In all settings the number of floors, i.e. prior hypotheses was set to 12; (b) Normalized loss along the depth of a belief tree in Case 2 with \budget$=3$ components and a planning horizon of $n=2$; (c) Normalized loss as a function of the size of the budget in Case 4, i.e. considering every valid heuristic in inference $h^{p\_candidate}$.  When \budget$\leq6$, the heuristic $h^{p\star}$ induces a smaller loss than $h^{inf}$. When \budget$>6$, both $h^{p\star}$ and $h^{inf}$ induce zero loss, i.e. are optimal in this setting.}
	\label{fig: main results with budgets}
	\vspace{-16pt}
\end{figure*}

In our second experiment we consider Case 2. In Fig. \ref{fig: results scenario2} we present the loss as a function of the budget size. As expected, with higher budget constraints the loss in solution quality becomes smaller. Moreover, as can be seen in Fig. \ref{fig: results scenario2b} the loss is higher closer to the root of the belief tree, as bounds are accumulated in the non-myopic setting, increasing the overlap.

Considering Case 3, our experiments did not show any computational improvements between calculating the original cost function and using our approach. We indicate that this is because there is no exponential growth in the number of hypotheses within the horizon and our considered cost function is linear w.r.t.~the number of components. However, as seen in \cite{Sztyglic21arxiv}, using a different cost, which is beyond the scope of this work, our approach can reduce the computational complexity while providing guarantees in Case 3 as well.

Finally, we consider Case 4. We first report that under this setting the computational complexity is high as every possible heuristic under the given budget is considered. In Fig. \ref{fig: results scenario4}   preliminary results indicate that this process can improve the bounds over the loss in solution quality vs a given heuristic $h^{inf}$.

% conclusions
\section{Conclusions}
In this work we introduced ND2A-BSP, an approach to reduce the computational complexity in data association aware BSP with performance guarantees for the nonmyopic case. We rigorously analyzed our approach considering different settings under budget constraints in inference and/or planning. 

Furthermore, future research will consider how to utilize information from planning in inference
when the latter is subject to hard computational budget constraints, as in most real-world autonomous systems.

\bibliographystyle{splncs03}
\bibliography{../../references/refs}

%\begin{thebibliography}{6}
%

%\bibitem {smit:wat}
%Smith, T.F., Waterman, M.S.: Identification of common molecular %subsequences.
%J. Mol. Biol. 147, 195?197 (1981). \url{doi:10.1016/0022-2836(81)90087-5}

%\bibitem {may:ehr:stein}
%May, P., Ehrlich, H.-C., Steinke, T.: ZIB structure prediction %pipeline:
%composing a complex biological workflow through web services.
%In: Nagel, W.E., Walter, W.V., Lehner, W. (eds.) Euro-Par 2006.
%LNCS, vol. 4128, pp. 1148?1158. Springer, Heidelberg (2006).
%\url{doi:10.1007/11823285_121}

%\bibitem {fost:kes}
%Foster, I., Kesselman, C.: The Grid: Blueprint for a New Computing %Infrastructure.
%Morgan Kaufmann, San Francisco (1999)

%\bibitem {czaj:fitz}
%Czajkowski, K., Fitzgerald, S., Foster, I., Kesselman, C.: Grid %information services
%for distributed resource sharing. In: 10th IEEE International %Symposium
%on High Performance Distributed Computing, pp. 181?184. IEEE Press, %New York (2001).
%\url{doi: 10.1109/HPDC.2001.945188}

%\bibitem {fo:kes:nic:tue}
%Foster, I., Kesselman, C., Nick, J., Tuecke, S.: The physiology of %the grid: an open grid services architecture for distributed systems %integration. Technical report, Global Grid
%Forum (2002)

%\bibitem {onlyurl}
%National Center for Biotechnology Information. %\url{http://www.ncbi.nlm.nih.gov}

%\end{thebibliography}
\end{document}